\documentclass[review]{siamart1116}

\usepackage{lipsum}
\usepackage{amsfonts}
\usepackage{graphicx}
\usepackage{epstopdf}
\usepackage{algorithmic}
\ifpdf
  \DeclareGraphicsExtensions{.eps,.pdf,.png,.jpg}
\else
  \DeclareGraphicsExtensions{.eps}
\fi

\usepackage{csquotes}

\usepackage{nicefrac}

\numberwithin{theorem}{section}

\newcommand{\TheTitle}{Interacting particles with Levy strategies} 
\newcommand{\TheAuthors}{G. Estrada-Rodriguez and H. Gimperlein}

\headers{\TheTitle}{\TheAuthors}

\title{{Interacting particles with L\'{e}vy strategies: limits of transport equations for swarm robotic systems}\thanks{We thank  Jose A.~Carrillo, Siobhan Duncan, Kevin J.~Painter, Jakub Sto\v{c}ek and Patricia A.~Vargas for fruitful discussions. Jakub Sto\v{c}ek provided generous assistance with the numerical experiments. }
\funding{G.~E.~R.~was supported by The Maxwell Institute Graduate School in Analysis and its
Applications, a Centre for Doctoral Training funded by the UK Engineering and Physical
Sciences Research Council (grant EP/L016508/01), the Scottish Funding Council, Heriot-Watt
University and the University of Edinburgh. }}

\author{
  Gissell Estrada-Rodriguez\thanks{Maxwell Institute for Mathematical Sciences and Department of Mathematics, Heriot-–Watt University, Edinburgh, EH14 4AS, United Kingdom
    (\email{ge5@hw.ac.uk}, \email{h.gimperlein@hw.ac.uk}).}
  \and
  Heiko Gimperlein${}^\dag$
}

\usepackage{amsopn}

\ifpdf
\hypersetup{
  pdftitle={\TheTitle},
  pdfauthor={\TheAuthors}
}
\fi

\nolinenumbers
\usepackage{braket,amsfonts}

\usepackage{mathtools}

\usepackage{amsfonts}
\usepackage{dsfont}

\usepackage{soul}

\makeatletter
\newcommand*{\rom}[1]{\expandafter\@slowromancap\romannumeral #1@}
\makeatother

\usepackage{array}

\usepackage[caption=false]{subfig}

\newsiamthm{claim}{Claim}
\newsiamremark{rem}{Remark}
\newsiamremark{expl}{Example}
\newsiamremark{hypothesis}{Hypothesis}
\crefname{hypothesis}{Hypothesis}{Hypotheses}

\usepackage{algorithmic}

\usepackage{graphicx,epstopdf}

\Crefname{ALC@unique}{Line}{Lines}

\usepackage{amsopn}

\numberwithin{theorem}{section}

\usepackage{xspace}
\usepackage{bold-extra}
\usepackage[most]{tcolorbox}
\definecolor{OliveGreen}{rgb}{0,0.6,0}

\colorlet{texcscolor}{blue!50!black}
\colorlet{texemcolor}{red!70!black}
\colorlet{texpreamble}{red!70!black}
\colorlet{codebackground}{black!25!white!25}

\lstdefinestyle{siamlatex}{%
  style=tcblatex,
  texcsstyle=*\color{texcscolor},
  texcsstyle=[2]\color{texemcolor},
  keywordstyle=[2]\color{texemcolor},
  moretexcs={cref,Cref,maketitle,mathcal,text,headers,email,url},
}

\tcbset{%
  colframe=black!75!white!75,
  coltitle=white,
  colback=codebackground, 
  colbacklower=white, 
  fonttitle=\bfseries,
  arc=0pt,outer arc=0pt,
  top=1pt,bottom=1pt,left=1mm,right=1mm,middle=1mm,boxsep=1mm,
  leftrule=0.3mm,rightrule=0.3mm,toprule=0.3mm,bottomrule=0.3mm,
  listing options={style=siamlatex}
}

\newtcblisting[use counter=example]{example}[2][]{%
  title={Example~\thetcbcounter: #2},#1}

\newtcbinputlisting[use counter=example]{\examplefile}[3][]{%
  title={Example~\thetcbcounter: #2},listing file={#3},#1}

\DeclareTotalTCBox{\code}{ v O{} }
{ 
  fontupper=\ttfamily\color{black},
  nobeforeafter,
  tcbox raise base,
  colback=codebackground,colframe=white,
  top=0pt,bottom=0pt,left=0mm,right=0mm,
  leftrule=0pt,rightrule=0pt,toprule=0mm,bottomrule=0mm,
  boxsep=0.5mm,
  #2}{#1}

\patchcmd\newpage{\vfil}{}{}{}
\begin{document}

\maketitle

\begin{abstract}
L\'{e}vy robotic systems combine superdiffusive random movement with emergent collective behaviour from local communication and alignment in order to find rare targets or track objects. In this article we derive macroscopic fractional PDE descriptions from the movement strategies of the individual robots. Starting from a kinetic equation which describes the movement of robots based on alignment, collisions and occasional long distance runs according to a L\'{e}vy distribution, we obtain a system of evolution equations for the fractional diffusion for long times. We show that the system allows efficient parameter studies for a  search problem, addressing basic questions like the optimal number of robots needed to cover an area in a certain time.  For shorter times, in the hyperbolic limit of the kinetic equation, the PDE model is dominated by alignment, irrespective of the long range movement. This is in agreement with previous results in swarming of self-propelled particles. The article indicates the novel and quantitative modeling opportunities which swarm robotic systems provide for the study of both emergent collective behaviour and anomalous diffusion, on the respective time scales.

\end{abstract}



\section{Introduction}\label{sec: introduction}
The automated searching of an area for a rare target and tracking are problems of long history in different areas of computer science \cite{couceiro2014benchmark}. They include search and rescue operations in disaster regions \cite{kantor2003distributed}, exploration for natural resources, environmental monitoring \cite{varela2011swarm} and surveillance. Systems of mobile robots have inherent advantages for these applications, compared to a single robot: the parallel and spatially distributed execution of tasks gives rise to larger sensing capabilities and efficient, fault tolerant strategies. See \cite{tan2013research} for a review \textcolor{black}{of} recent advances on swarm robotics and applications.

In this article we consider macroscopic PDE descriptions applicable to swarm robotic systems, which achieve scalability for a large number of independent, simple robots based on  local communication and emergent collective behaviour. Much of the research  focuses on determining control laws of the robot movement which give rise to a desired group behavior \cite{brambilla}, like a prescribed spatial distribution. Typical control laws like biased random walks, reaction to chemotactic cues and long range coordination, are reminiscent of models for biological systems, and many bio-inspired strategies have been implemented in robots in recent years, for a review see \cite{ senanayake2016search}.


Of particular recent interest have been strategies which include \emph{nonlocal} random  movements beyond Brownian motion, leading to L\'{e}vy robotics \cite{krivonosov2016vy}. L\'{e}vy walks, with the characteristic high number of long runs, minimize the expected hitting time to reach an unknown target. These new search strategies are inspired by nonlocal movement found in a variety of organisms like T cells \cite{harris}, \textit{E.~coli} bacteria \cite{korobkova2004molecular}, mussels \cite{de2011levy} and spider monkeys \cite{ramos2004levy}. Conversely, robotic systems provide controlled, quantitative models rarely available in biology.

Given sets of control laws are assessed and optimized by expensive particle based simulations and experiments with robots, based on a wide range of quality metrics \cite{andersonquantitative,brambilla,zhang2018performance}. On the other hand, for biological systems effective macroscopic PDE descriptions have proven to be a key tool for  efficient parameter optimization and analytical understanding. A series of studies dating to Patlak \cite{patlak1953random} has generated solid
understanding on how microscopic detail translates into a diffusion-advection
type equation \cite{bellomo2012mathematical,xue2009individual} for random walks subject to an external bias and interactions. Recent work has made progress towards nonlocal PDE descriptions of  L\'{e}vy movement \cite{pks2018,perthame2018fractional,taylor2016fractional}. Emergence of superdiffusion without L\'{e}vy movement is discussed in \cite{fedotov2017emergence}.

 In this article, motivated by the necessity of optimal search strategies for a swarm of robots, we study a system of $N$ individuals undergoing a velocity jump process with contact interactions and where the individuals align with their neighbors. \textcolor{black}{We assume  quasi-static behavior, i.e., along a single run  changes in time of the robotic parameters are negligible, and use the molecular chaos assumption, that the velocity of particles which are about to collide is approximately uncorrelated. In a mean-field limit for $N \to \infty$,} we obtain a system of fractional PDEs for the macroscopic density
$u(\mathbf{x},t)$ and mean direction $w(\mathbf{x},t)$:
\begin{equation}
 \begin{aligned}
     \partial_tu+\nabla\cdot w & =0\ ,  \\ w-\ell\frac{G(u)}{F(u)}\Lambda^w & =-\frac{1}{F(u)}C_\alpha\nabla^{\alpha-1}u\ ,\label{eq: 1}
 \end{aligned}
 \end{equation}\label{ourmaineq}
 both assumed to be sufficiently smooth and integrable in $\mathds{R}^n$.
Here $\Lambda^w$ is given by \cref{eq: new alignment} and $F(u)$, $G(u)$ and the diffusion constant $C_\alpha$ are defined in defined by \cref{eq: F y G} and \cref{eq: diffusion constant} respectively. The parameter $\ell$ gives the strength of the alignment.
Starting from a kinetic equation that describes the movement of the individuals, combining short range interactions and alignment with occasional long runs, according to an approximate L\'{e}vy distribution, we obtain the system \cref{eq: 1} in the appropriate parabolic time scale.

While diffusive behavior dominates for long times, swarming on shorter hyperbolic time scales is not affected by the L\'{e}vy movement, so that a rich body of work such as \cite{bostan2017reduced} on swarming applies to L\'{e}vy robotics. Combining long range dispersal and alignment as in the kinetic equation for the macroscopic density \cref{eq: similar to previous paper}, allows us to obtain either a space fractional diffusion equation for a pure nonlocal movement of the individuals (see \cite{pks2018}), or a Vicsek-type equation for the case of pure alignment \cite{degond2004macroscopic,vicsek1995novel}.

To illustrate applications of the PDE description, Section \ref{sec: computation of relevant quantities} presents efficient numerical methods for the solution of \cref{eq: 1} and applies them for some first parameter studies in the case of \textcolor{black}{a search example.} Detailed robotic studies of \textcolor{black}{search strategies and targeting efficiency}, \textcolor{black}{based on \eqref{ourmaineq},} as well as comparisons to both standard particle simulations and experiments with \textit{E-Puck} robots and drones \textcolor{black}{are pursued in \cite{duncanestrada}}.

Concerning previous experimental work, the particular L\'{e}vy strategy considered here, with additional long waiting times during reorientations, was implemented in a swarm robotic system of \textit{iAnt} robots to find targets in \cite{fricke2016immune}, while \cite{sutantyo2010multi} combined a L\'{e}vy walk search strategy with an added repulsion among the robots. L\'{e}vy movement directed by external cues, such as chemotaxis, has been studied in \cite{pasternak2009levy} to find a contaminant in water, while \cite{nurzaman2009yuragi} considers sonotaxis. Efficient spatial coverage in the presence of pheromone cues was specifically addressed in \cite{schroeder2017efficient}, using an ant-inspired search strategy with long range movement. \\

\noindent \emph{Notation:} The words \emph{particles} and \emph{individuals} are used interchangibly in this work. We denote the unit sphere in $\mathbb{R}^n$ by $S = \{x\in \mathbb{R}^n : |x|=1\}$, its surface area by $|S|$.

\section{Model assumptions}\label{sec: model}
A swarm robotic system \cite{senanayake2016search} consists of a large number of simple independent robots with local rules, communication and interactions among them and with the environment, where the local interactions may lead to collective behaviour of the swarm. A system of  \textit{E-Puck} robots \cite{epuck} in a domain in $\mathbb{R}^2$ provides a specific model system, to which we apply our results and present numerical experiments in Section \ref{sec: computation of relevant quantities}. More generally, we consider $N$ identical spherical individuals of diameter $\varrho>0$ in $\mathds{R}^n$, where $n=2,3$. Each individual is characterized by its position $\mathbf{x}_i\in\mathds{R}^n$ and direction $\mathbf{\theta}_i\in S=\{|\mathbf{x}_i|=1 \}\subseteq\mathds{R}^n$. We assume that each individual moves according to the following rules:
\begin{enumerate}
\item Starting at position $\mathbf{x}$ at time $t$, an individual runs in
direction $\theta$ for a L\'{e}vy distributed time $\tau$, called the \enquote{run time}.
\item The individuals move according to a velocity jump process with constant forward speed $c$, following a straight line motion interrupted by reorientation.
\item When the individual stops, with probability $\zeta$ it starts a long range run and tumble process, choosing a new direction $\theta^*$ according to a distribution $k(\mathbf{x},t,\mathbf{\theta};\mathbf{\theta}^*)$. With probability $(1-\zeta)$ it aligns with the neighbors in a certain region.
\item When two individuals get close to each other they reflect elastically; the new direction is $\theta'=\theta-2(\theta\cdot\nu)\nu$, where $\nu=\frac{\mathbf{x}_i-\mathbf{x}_j}{|\mathbf{x}_i-\mathbf{x}_j|}$ is the normal vector at the point of collision.
\item All reorientations are assumed to be instantaneous.
\item The running\footnote{In probability this is also known as survival probability, where the \enquote{event} in this case is to stop. Hence \enquote{survival} in that context refers to the probability of continuing to move in the same direction for some time $\tau$.} probability $\psi$, which is defined as the probability that an individual moving in some fixed direction does not stop until time $\tau$, is taken to be independent on the environment surrounding the individual.
\end{enumerate}

Note that the assumptions correspond to independent individuals with simple capabilities relative to typical tasks for swarm robotic systems. They interact only with their neighbors in a fixed sensing region, and the movement decisions are based on the current positions and velocities, not information from earlier times.  This assures the scalability to large numbers of robots, while nonlocal collective movement may emerge from the local rules \cite{senanayake2016search}.

Related movement laws have been used for target search, for example, in the experiments in \cite{fricke2016immune}.  Refined local control laws and the possibility for quantitative experiments with robots open up novel modeling opportunities, see \Cref{sec: discussion}.
\section{Kinetic equation for microscopic movement}\label{sec: collision description}

For the system of $N$ individuals \textcolor{black}{moving on the trajectories $\{\mathbf{X}_i(t)\}_{i=1}^N$} described in \Cref{sec: model}, \textcolor{black}{let us denote by $\sigma= \sigma(\mathbf{x}_i,t,\theta_i,\tau_i)$ the $N$-particle probability density function. This means that the probability density of finding the individuals at time $t$ at position $\mathbf{x}_i$ moving in direction $\theta_i$ with run time $\tau_i$ is given by $\sigma(\mathbf{x}_i,t,\theta_i,\tau_i) \prod_{i=1}^N d\mathbf{x}_i\ d\theta_i\ d\tau_i$.
At least formally one expects that $\sigma$ evolves according to the kinetic equation given by} \cite{kennard1938kinetic}
\begin{equation}
    \partial_t\sigma+\sum_{i=1}^N\left(\partial_{\tau_i}+\textcolor{black}{c}\theta_i\cdot\nabla_{\mathbf{x}_i}\right)\sigma=-\sum_{i=1}^N\beta_i\sigma\ ,\label{eq: N particles}
\end{equation}
\textcolor{black}{in the domain $\Omega^N = \{(\mathbf{x}_1, ..., \mathbf{x}_N) \in \mathds{R}^{n\times N}:\ |\mathbf{x}_i-\mathbf{x}_j|\geq\varrho \ \forall i,j \}$}. 
\textcolor{black}{The tumbling \cref{eq: turn angle operator} and the alignment \cref{eq: alignment expressions} described below determine the initial condition for the kinetic equation \cref{eq: N particles} at $\tau_i=0$ in  \cref{eq: final new directionnew} such that  mass is conserved. The equation is complemented by an initial condition at time $t=0$ (smooth, compactly supported) and boundary conditions at $\partial \Omega^N$ corresponding to elastic collisions.}

The stopping frequency $\beta_i$ during a run phase  relates to the probability $\psi_i$ that an individual does not stop for a time $\tau_i$. It is given by
\begin{equation}
\psi_i(\mathbf{x}_i,\tau_i)=\left(\frac{\textcolor{black}{\varsigma_0(\mathbf{x}_i)}}{\textcolor{black}{\varsigma_0(\mathbf{x}_i})+\tau_i}\right)^{\alpha},\ \alpha\in(1,2)\ . \label{eq: survival}
\end{equation}
This power law behaviour corresponds to the long tailed distribution of run times described in Assumption 1.~in \Cref{sec: model}, instead of the Poisson process in classical velocity jump models \cite{erban2004individual, othmer2000diffusion, othmer2002diffusion}. As the speed $c$ of the runs is constant, the individuals perform occasional long jumps with a power-law distribution of run
lengths. The stopping frequency is given by
\begin{equation}
    \beta_i(\mathbf{x}_i,\tau_i)=-\frac{\partial_{\tau_i}\psi_i}{\psi_i}=\frac{\varphi_i}{\psi_i}\ .\label{eq: beta}
\end{equation}

After stopping, according to Assumption 3~individuals choose a new direction of motion by either tumbling or alignment. With probability $\zeta \in [0,1]$ they
choose a new direction according to the turning kernel $T_i$ given by
\begin{equation}
    T_i\phi(\theta_i^*)=\int_S k(\mathbf{x}_i,t,\mathbf{\theta}_i;\mathbf{\theta}_i^*)\phi(\theta_i)d\theta_i \label{eq: turn angle operator}\ ,
\end{equation}
where the new direction $\theta_i^*$ is symmetrically distributed with respect to the previous
direction $\theta_i$ according to the distribution $k(\mathbf{x}_i,t,\mathbf{\theta}_i;\mathbf{\theta}_i^*)=\tilde{k}(\mathbf{x}_i,t,|\theta_i^*-\theta_i|)$  \cite{alt1980biased}.
\textcolor{black}{Because $\tilde{k}$ is a probability distribution, it is normalized to $\int_S\tilde{k}(\mathbf{x}_i,t,|\theta_i-e_1|)d\theta_i=1$
where $e_1 = (1, 0, . . . , 0)$. }

With probability $(1-\zeta)$ the new direction of motion is aligned with the direction of the neighbors according to a distribution $\Phi(\Lambda_i\cdot\theta_i)$, with
$\int_S\Phi(\Lambda_i\cdot\theta_i)d\theta_i=1$. The average direction $\Lambda_i$ at $\mathbf{x}_i$  is defined in terms of the nonlocal flux $\mathcal{J}(\mathbf{x}_i)$ \cite{dimarco2016self}, 
\begin{equation}
\Lambda_i(\mathbf{x}_i,\theta_i,t)=\frac{\mathcal{J}(\mathbf{x}_i,t)}{|\mathcal{J}(\mathbf{x}_i,t)|}\ ,\ \mathcal{J}(\mathbf{x}_i,t)=\int_{\mathds{R}^n}\int_SK(|\mathbf{x}_j-\mathbf{x}_i|)p(\mathbf{x}_j,t,\theta_j)\theta_jd\mathbf{x}_jd\theta_j \ .\label{eq: alignment expressions}
\end{equation}
Here $K$ is a given influence kernel and $p$ is the density of individuals at $\mathbf{x}_j$ at time $t$, moving in the direction $\theta_j$ defined as (for $j=1$)
\begin{align*}
&    p(\mathbf{x}_1,t,\theta_1)=\\ & \ \ \frac{1}{|S|^N}\int_{[0,t]^N}\int_{\Omega_{N-1}(\mathbf{x}_1)}\int_{S^N}\sigma(\mathbf{x}_1,\theta_1,\tau_1, \dots,\mathbf{x}_N,\theta_N,\tau_N,t) d\theta_{2}d\mathbf{x}_{2}d\tau_{2}\dots d\theta_{N}d\mathbf{x}_{N}d\tau_{N}\ .
\end{align*}
\textcolor{black}{The integral is over the domain $\Omega_{N-1}(\mathbf{x}_1) := \{(\mathbf{x}_2,\dots, \mathbf{x}_N): (\mathbf{x}_1, \mathbf{x}_2,\dots, \mathbf{x}_N) \in \Omega^{N}\}$ accessible to individuals $2, \dots, N$.} \textcolor{black}{If the flux $\mathcal{J}(\mathbf{x}_i,t)=0$, we assume then that $\Lambda_i(\mathbf{x}_i,\theta_i,t)$ takes the value $\theta_i$ \cite{degond2008continuum}. In the rest of the paper this convention will be recalled by the dependence of $\Lambda_i$ on $\theta_i$.}

\section{Transport equation for the two-particle density}\label{sec: two particle case}

\textcolor{black}{In the following we assume that  $\sigma$ and its derivatives are smooth. As the initial condition at time $0$ is of compact support, $\sigma$ is of compact support for every fixed $t$.} The description \cref{eq: N particles} of the $N$-particle problem a priori requires the understanding of collisions among the whole system of particles. In this article, however, we aim for a macroscopic description for low densities, as made precise by the scaling in \Cref{sec: scaling}. In this \textcolor{black}{regime} collisions of more than two individuals \textcolor{black}{are} neglected \cite{cercignani2013mathematical}, and we truncate the hierarchy of equations by neglecting collisions of $3$ or more individuals and integrate out individuals $3,...,N$ from $\sigma$.  The transport equation which describes the movement of two particles \textcolor{black}{$\mathbf{x}_1,\ \mathbf{x}_2\in \Omega^2$}, is given by 
\begin{align}
    \partial_{\tau_1}\sigma+\partial_{\tau_2}\sigma+\partial_t\sigma+c\theta_1\cdot\nabla_{\mathbf{x}_1}\sigma+c\theta_2\cdot\nabla_{\mathbf{x}_2}\sigma=-(\beta_1+\beta_2)\sigma \label{eq: initial model}\ .
\end{align}
Here $\sigma=\sigma(\mathbf{x}_1,\mathbf{x}_2,t,\theta_1,\theta_2,\tau_1,\tau_2)$ is the two-particle density function \textcolor{black}{and we impose the boundary and initial conditions as for \cref{eq: N particles}.} 

We first integrate with respect to $\tau_1$ and $\tau_2$ to get 
\begin{align}
    \partial_t\tilde{\tilde{\sigma}}+c\theta_1\cdot\nabla_{\mathbf{x}_1}\tilde{\tilde{\sigma}}&+c\theta_2\cdot\nabla_{\mathbf{x}_2}\tilde{\tilde{\sigma}} =  -{ \int_0^t\int_0^t\beta_1\sigma d\tau_1d\tau_2}  { -\int_0^t\int_0^t\beta_2\sigma d\tau_1d\tau_2}\nonumber\\ &\textcolor{black}{+\tilde{\sigma}_{\tau_1}(\mathbf{x}_1,\mathbf{x}_2,t,\theta_1,\theta_2,\tau_1=0)+\tilde{\sigma}_{\tau_2}(\mathbf{x}_1,\mathbf{x}_2,t,\theta_1,\theta_2,\tau_2=0)} \ ,\label{eq: tau independent equation}
\end{align}
for
\[
\tilde{\tilde{\sigma}}(\mathbf{x}_1,\mathbf{x}_2,t,\theta_1,\theta_2)= \int_0^t\int_0^t\sigma d\tau_1d\tau_2\ ,\ \ \tilde{\sigma}_{\tau_1}(\mathbf{x}_1,\mathbf{x}_2,t,\theta_1,\theta_2,\tau_{1})= \int_0^t\sigma d\tau_2\ ,\]
\textcolor{black}{and similarly for $\tilde{\sigma}_{\tau_2}$. The last two terms in \Cref{eq: tau independent equation} come from applying the Fundamental Theorem of Calculus to $\int_0^t\partial_{\tau_1}\sigma d\tau_1$ and $\int_0^t\partial_{\tau_2}\sigma d\tau_2$.} 

After stopping with rate given by $\beta_1$, from \Cref{sec: collision description}, the initial condition for the new run of individual $1$ is given by  
\begin{equation}\label{eq: final new directionnew}
\tilde{\sigma}_{\tau_1}(\mathbf{x}_1,\mathbf{x}_2,t,\theta_1,\theta_2,\textcolor{black}{\tau_1=0}) =\int_SQ(\theta_1,\theta_1^*)\int_0^t\beta_1\tilde{\sigma}_{\tau_1}(\mathbf{x}_1,\mathbf{x}_2,t,\theta_1^*,\theta_2,\tau_1)d\tau_1d\theta_1^*\  ,
\end{equation}
where
\begin{equation}
Q(\theta_1,\theta_1^*)=\zeta k(\mathbf{x}_1,t,\theta_1^*;\theta_1)+(1-\zeta)\Phi(\Lambda_1\cdot\theta_1)\ .
\end{equation}
\textcolor{black}{The operator $Q(\theta_1,\theta_1^*)$ satisfies Assumption 3. from \Cref{sec: model}.} In absence of collisions and  for $\zeta=0$ we recover the kinetic equation for alignment interactions as in \cite{degond2013macroscopic,dimarco2016self,ha2008particle}, while for $\zeta=1$ we obtain the long range velocity jump process from \cite{pks2018}.
Substituting \cref{eq: final new directionnew} and its analogue for individual 2 into the kinetic equation for $\tilde{\tilde{\sigma}}$, we obtain

\begin{equation}
\begin{aligned}
\underset{(\textnormal{\rom{1}})}{\underbrace{\partial_{t}\tilde{\tilde{\sigma}}}}&+\underset{(\textnormal{\rom{2}})}{\underbrace{c\theta_{1}\cdot\nabla_{\mathbf{x}_{1}}\tilde{\tilde{\sigma}}}}+\underset{(\textnormal{\rom{3}})}{\underbrace{c\theta_{2}\cdot\nabla_{\mathbf{x}_{2}}\tilde{\tilde{\sigma}}}}=-\underset{(\textnormal{\rom{4}})}{\underbrace{(\mathds{1}-\zeta T_1)\int_0^t\int_0^t\beta_{1}\sigma d\tau_1d\tau_2}} \\&-\underset{(\textnormal{\rom{5}})}{\underbrace{(\mathds{1}-\zeta T_2)\int_0^t\int_0^t\beta_2\sigma d\tau_1d\tau_2}}+\underset{(\textnormal{\rom{6}})}{\underbrace{(1-\zeta)|S|\Phi(\Lambda_1\cdot\theta_1)\int_0^t\beta_1P_1d\tau_1}}\\&+\underset{(\textnormal{\rom{7}})}{\underbrace{(1-\zeta)|S|\Phi(\Lambda_2\cdot\theta_2)\int_0^t\beta_2P_2d\tau_2}}\ .\label{eq: system}
\end{aligned}
\end{equation}
where
\[
P_1(\mathbf{x}_1,\mathbf{x}_2,t,\theta_2,\tau_1)=\frac{1}{|S|}\int_S\tilde{\sigma}_{\tau_1}(\mathbf{x}_1,\mathbf{x}_2,t,\theta_1^*,\theta_2,\tau_1)d\theta_1^*
\]
and $P_2$ is similarly defined. Here $|S|$ denotes the surface area of the unit sphere $S$.

From the method of characteristics, we note that the solution of \cref{eq: initial model} is
\begin{equation}
    \sigma=\sigma_0(\mathbf{x}_1-c\theta_1\tau_1,\mathbf{x}_2-c\theta_2\tau_1,t-\tau_1,\theta_1,\theta_2,0,\tau_2-\tau_1)\psi_1(\mathbf{x}_1,\tau_1)\frac{\psi_2(\mathbf{x}_2,\tau_2)}{\psi_2(\mathbf{x}_2,\tau_2-\tau_1)}\ \label{eq: method of characteristics}.
\end{equation}

\begin{figure}
    \centering
   \includegraphics[scale=0.4]{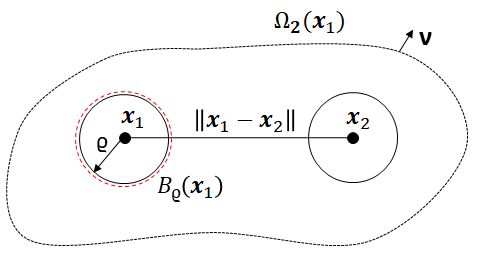}
    \caption{Illustration of the collision domain between two individuals.}
    \label{fig: domain}
\end{figure}

From equation \cref{eq: system} for the two-particle density function $\tilde{\tilde{\sigma}}$ we now aim to derive an effective transport equation for the one-particle density function
\begin{equation}
    p(\mathbf{x}_1,t,\theta_1)=\frac{1}{|S|}\int_0^t\int_0^t\int_{\Omega_2}\int_{S}\sigma d\theta_2d\mathbf{x}_2d\tau_1d\tau_2\ .\label{eq: density of 1}
\end{equation}
By integrating equation \cref{eq: system} with respect to the accessible phase space $(\mathbf{x}_2, \theta_2)\in\Omega_2\times S$, where
$
\Omega_2=\Omega_2(\mathbf{x}_1)=\{\mathbf{x}_2\in\mathds{R}^n:\ |\mathbf{x}_1-\mathbf{x}_2|>\varrho  \} = \mathbb{R}^n\setminus B_\varrho(\mathbf{x}_1)
$
as in \Cref{fig: domain}, we obtain the following terms:

\vspace{0.1cm}

\noindent(\rom{1}) From our assumptions we commute the integrals and the time derivative, resulting in 
\[\int_{\Omega_2}\int_{S}\partial_t\tilde{\tilde{\sigma}}d\theta_2d\mathbf{x}_2=|S|\partial_tp\ .\]

\vspace{0.1cm}

\noindent(\rom{2}) From Reynolds' \textcolor{black}{transport theorem in the variable $\mathbf{x}_1$}
\begin{equation*}
    c\int_{\Omega_2}\int_{S}\theta_1\cdot\nabla_{\mathbf{x}_1}\tilde{\tilde{\sigma}}d\theta_2d\mathbf{x}_2=|S|c\theta_1\cdot\nabla_{\mathbf{x}_1}p-c\int_{\partial B_\varrho(\mathbf{x}_1)}\int_{S}(\theta_1\cdot\nu)\tilde{\tilde{\sigma}}d\theta_2d\mathbf{x}_2\ .
\end{equation*}
$\nu$ is the outward pointing unit normal vector with respect to $\Omega_2$.

\vspace{0.1cm}

\noindent(\rom{3}) From the divergence theorem
\begin{align*}
    c\int_{\Omega_2}\int_{S}\theta_2\cdot\nabla_{\mathbf{x}_2}\tilde{\tilde{\sigma}}d\theta_2d\mathbf{x}_2 &  =c\int_{\partial B_\varrho}\int_S(\theta_2\cdot\nu)\tilde{\tilde{\sigma}}d\theta_2d\mathbf{x}_2\ ,
\end{align*}
as $\tilde{\tilde{\sigma}}$ has compact support.
\vspace{0.1cm}

\noindent(\rom{4}) Changing the order of integration we have
\begin{align*}
    (\mathds{1}-\zeta T_1)& \int_{\Omega_2}\int_{S}\int_0^t\int_0^t\beta_1(\mathbf{x}_1,\tau_1)\sigma(\mathbf{x}_1,\mathbf{x}_2,t,\theta_1,\theta_2,\tau_1,\tau_2) d\tau_1d\tau_2d\theta_2d\mathbf{x}_2\\ & =|S|(\mathds{1}-\zeta T_1)\int_0^t\beta_1(\mathbf{x}_1,\tau_1)J_1(\mathbf{x}_1,t,\theta_1,\tau_1)d\tau_1\ ,
\end{align*}
for  $J_1(\mathbf{x}_1,t,\theta_1,\tau_1)=\int_{\Omega_2}\int_S\int_0^t\sigma(\mathbf{x}_1,\mathbf{x}_2,t,\theta_1,\theta_2,\tau_1,\tau_2) d\tau_2d\theta_2d\mathbf{x}_2$.

\vspace{0.1cm}
\noindent(\rom{5}) Similar to (\rom{4}) and using 
\[
\int_ST\phi(\cdot,\theta)d\theta =\int_S\phi(\cdot,\eta)\int_Sk(\cdot,\eta;\theta)d\theta d\eta=\int_S\phi(\cdot,\eta)d\eta=\phi(\cdot)\ ,
\]
where $\phi$ is an arbitrary function, we obtain
\begin{align*}
    \int_{\Omega_2}\int_{S}(\mathds{1}&-\zeta T_2) \int_0^t\int_0^t\beta_2(\mathbf{x}_2,\tau_2)\sigma(\mathbf{x}_1,\mathbf{x}_2,t,\theta_1,\theta_2,\tau_1,\tau_2)d\tau_1d\tau_2\theta_2d\mathbf{x}_2\\ & =|S|(1-\zeta)\int_{\Omega_2}\int_0^t\beta_2(\mathbf{x}_2,\tau_2)P_2(\mathbf{x}_1,\mathbf{x}_2,t,\theta_1,\tau_2)d\tau_2d\mathbf{x}_2\ .
\end{align*}

 \vspace{0.1cm}
\noindent(\rom{6}) Moreover,
\begin{align*}
(1-\zeta)|S|\Phi(\Lambda_1\cdot\theta_1) & \int_S\int_{\Omega_2}\int_0^t\beta_1(\mathbf{x}_1,\tau_1)P_1(\mathbf{x}_1,\mathbf{x}_2,t,\theta_2,\tau_1)d\tau_1d\mathbf{x}_2d\theta_2\nonumber\\ &= |S|^2(1-\zeta)\Phi(\Lambda_1\cdot\theta_1)\int_0^t\beta_1(\mathbf{x}_1,\tau_1)\bar{P}_1(\mathbf{x}_1,t,\tau_1)d\tau_1 \ ,
\end{align*}
where $
\bar{P}_1(\mathbf{x}_1,t,\tau_1)=|S|^{-1}\int_{\Omega_2}\int_S P_1(\mathbf{x}_1,\mathbf{x}_2,t,\theta_2,\tau_1)d\theta_2d\mathbf{x}_2$.

\vspace{0.1cm}
\noindent(\rom{7}) Recalling the normalization
$\int_S\Phi(\Lambda_2\cdot\theta_2)d\theta_2=1$, we conclude
\begin{align*}
(1-\zeta)|S|&\int_S\Phi(\Lambda_2\cdot\theta_2)\int_{\Omega_2}  \int_0^t\beta_2(\mathbf{x}_2,\tau_2)P_2(\mathbf{x}_1,\mathbf{x}_2,t,\theta_1,\tau_2)d\tau_2d\mathbf{x}_2d\theta_2\nonumber\\ &= |S|(1-\zeta)\int_{\Omega_2}\int_0^t\beta_2(\mathbf{x}_2,\tau_2)P_2(\mathbf{x}_1,\mathbf{x}_2,t,\theta_1,\tau_2)d\tau_2d\mathbf{x}_2\ . \nonumber
\end{align*}

From the above analysis we see that (\rom{5}) and (\rom{7}) cancel. The final step now is to write  (\rom{4}) and (\rom{6}) in terms of the density $p(\mathbf{x}_1,t,\theta_1)$ defined in \cref{eq: density of 1}. We proceed as follows. 

From (\rom{4}) and using the solution \cref{eq: method of characteristics} we can write the integral as
\begin{align}
    \int_0^t&\beta_1(\mathbf{x}_1,\tau_1)J_1(\mathbf{x}_1,t,\theta_1,\tau_1)d\tau_1\nonumber\\& = \int_0^t\int_0^t\varphi_1(\mathbf{x}_1,\tau_1)\bar{\sigma}_0(\mathbf{x}_1-c\theta_1\tau_1,t-\tau_1,\theta_1,0,\tau_2-\tau_1)\frac{\psi_2(\mathbf{x}_2,\tau_2)}{\psi_2(\mathbf{x}_2,\tau_2-\tau_1)}d\tau_1d\tau_2\nonumber\\ & =\int_0^t\varphi_1(\mathbf{x}_1,\tau_1)\bar{\bar{\sigma}}_0(\mathbf{x}_1-c\theta_1\tau_1,t-\tau_1,\theta_1,0)d\tau_1\ , \label{eq: new2}
\end{align}
 where we have used $\frac{\psi_2(\mathbf{x}_2,\tau_2)}{\psi_2(\mathbf{x}_2,\tau_2-\tau_1)}\rightarrow 1$ for $\tau_1\ll 1$, i.e. we assume the trajectories of particle one are very short in time. 
 From the definition \cref{eq: density of 1} and using \cref{eq: method of characteristics} again we write
 \begin{equation}
     p(\mathbf{x}_1,t,\theta_1)=\int_0^t\bar{\bar{\sigma}}_0(\mathbf{x}_1-c\theta_1\tau_1,t-\tau_1,\theta_1,0)\psi_1(\mathbf{x}_1,\tau_1)d\tau_1\ .\label{eq: new}
 \end{equation}
 The standard arguments used in \cite{pks2018}
 allow to write \cref{eq: new2} as a convolution by using the Laplace transform of \cref{eq: new2} and \cref{eq: new} as follows,
 \begin{equation}
     \int_0^t\beta_1(\mathbf{x}_1,\tau_1)J_1(\mathbf{x}_1,t,\theta_1,\tau_1)d\tau_1=\int_0^t\mathcal{B}(\mathbf{x}_1,t-s)p(\mathbf{x}_1-c\theta_1(t-s),s,\theta_1)ds.
 \end{equation}
Here the operator $\mathcal{B}$ is defined from its Laplace transform $\hat{\mathcal{B}}=\mathcal{L}\{\mathcal{B} \}$ in time,
\begin{equation}
    \hat{\mathcal{B}}(\mathbf{x}_1,\lambda+c\theta_1\cdot\nabla_{\mathbf{x}_1})=\frac{\hat{\varphi}_1(\mathbf{x}_1,\lambda+c\theta_1\cdot\nabla_{\mathbf{x}_1})}{\hat{\psi}_1(\mathbf{x}_1,\lambda+c\theta_1\cdot\nabla_{\mathbf{x}_1})}\ ,
\end{equation}
with $\varphi_1$ and $\psi_1$ from \cref{eq: beta}. Explicit expressions for $\hat{\varphi}_1$ and $\hat{\psi}_1$ are found below in \Cref{sec: fractional diffsuion equation}.

Following the same arguments we can write the integral in (\rom{6}) as follows
\begin{equation}
    \int_0^t\beta_1(\mathbf{x}_1,\tau_1)\bar{P}_1(\mathbf{x}_1,t,\tau_1)d\tau_1=\int_0^t\mathcal{B}(\mathbf{x}_1,t-s)u(\mathbf{x}_1,s)ds\ ,\label{eq: new3}
\end{equation}
where $u(\mathbf{x}_1,t)$ is a macroscopic density defined as 
\begin{equation}
    u(\mathbf{x}_1,t)=\int_Sp(\mathbf{x}_1,t,\theta_1)d\theta_1\ .\label{eq: u}
\end{equation}
Finally, including the results obtained in (\rom{1})-(\rom{3}) and the convolutions \cref{eq: new} and \cref{eq: new3} we obtain

\begin{equation}
\begin{aligned}
    \partial_tp +c\theta_1\cdot\nabla p& =c|S|^{-1}\int_{\partial B_\varrho}\int_{S}\nu\cdot(\theta_1-\theta_2)\tilde{\tilde{\sigma}}d\theta_2d\mathbf{x}_2\\ &\qquad +(1-\zeta)|S|\Phi(\Lambda_1\cdot\theta_1)\int_0^t\mathcal{B}(\mathbf{x}_1,t-s)u(\mathbf{x}_1,s)ds\\  &\qquad  -(\mathds{1}-\zeta T_1)\int_0^t\mathcal{B}(\mathbf{x}_1,t-s)p(\mathbf{x}_1-c\theta_1(t-s),s,\theta_1)ds\ .\label{eq: one particle equation}
\end{aligned}
\end{equation}

To summarize, the transport equation \cref{eq: one particle equation} describes the evolution of the one-particle density function $p(\mathbf{x}_1,t,\theta_1)$. The three terms on the right hand side describe the collisions, the alignment and the long range velocity jump process, respectively. \\


 For later convenience we rewrite the collision term
$
\int_{\partial B_\varrho}\int_S\nu\cdot(\theta_1-\theta_2)\tilde{\tilde{\sigma}}d\theta_2d\mathbf{x}_2
$ as in \Cref{sec: collision term}. Summing over the $N-1$ individuals which individual $1$ can collide with, equation \cref{eq: one particle equation} turns into
 \begin{equation}
 \begin{aligned}
 \partial_tp  &+c\theta_1\cdot\nabla_{\mathbf{x}_1}p=(1-\zeta)|S|\Phi(\Lambda_1\cdot\theta_1)\int_0^t\mathcal{B}(\mathbf{x}_1,t-s)u(\mathbf{x}_1,s)ds   \\&-(\mathds{1}-\zeta T_1)\int_0^t\mathcal{B}(\mathbf{x}_1,t-s)p(\mathbf{x}_1-c\theta_1(t-s),s,\theta_1)ds \\ &+|S|^{-1}c\varrho^{n-1}(N-1)\int_{S_+}\int_{S}\nu\cdot(\theta_1-\theta_2)\Bigl[\tilde{\tilde{\sigma}}(\mathbf{x}_1,\mathbf{x}_1-\nu\varrho,t,\theta_1',\theta_2')\\ &-\tilde{\tilde{\sigma}}(\mathbf{x}_1,\mathbf{x}_1+\nu\varrho,\theta_1,\theta_2) \Bigr]d\theta_2d\nu \ .  \label{eq: final one particle}
 \end{aligned}
 \end{equation}
 \textcolor{black} {Note that under the molecular chaos assumption each of the $N-1$ possible collision partners contributes an identical collision term to \eqref{eq: final one particle}, which is therefore multiplied by $N-1$.} From now on we work with equation \cref{eq: final one particle} which describes the evolution of the one-particle density $p$ in the system of $N$-particles.

 \section{Parabolic scaling}\label{sec: scaling}
In  applications, the  mean run time $\bar{\tau}$ is often small compared with the macroscopic time scale $\mathcal{T}$, and we aim to study \cref{eq: final one particle} for $\varepsilon=\nicefrac{\bar{\tau}}{\mathcal{T}}\ll 1$ \cite{alt1980biased}. Denoting a macroscopic length scale by $\mathcal{X}$ and $s=\frac{\mathcal{X}}{\mathcal{T}}$, we introduce normalized variables $$\bar{t}=\frac{t}{\mathcal{T}},\ \bar{\mathbf{x}}=\frac{\mathbf{x}}{\mathcal{X}},\ \bar{\tau}=\frac{\tau}{\bar{\tau}}\ \mathrm{and}\ \bar{c}=\frac{c}{s}\ .$$
A diffusion limit of \cref{eq: final one particle} is obtained under the scaling $(\mathbf{x},t,\tau)\mapsto(\bar{\mathbf{x}}s/\varepsilon,\thinspace \bar{t}/\varepsilon,\thinspace  \bar{\tau}/\varepsilon^\mu)$, with $\bar{c}=\varepsilon^{-\gamma}c_0$
for $\mu,\ \gamma>0$. We further assume that the diameter of each particle is small,
$\varrho=\varepsilon^\xi$, while the number of particles $N$ is large so that $(N-1)\varrho=\varepsilon^{\xi-\vartheta}$, with $\xi-\vartheta<0$. The scaling of the alignment is $\varepsilon^{-\eta}$.

In the normalized variables equations \cref{eq: beta} and \cref{eq: survival} become, after dropping the bar for the new variables,
\begin{equation}
\beta_\varepsilon(\mathbf{x}_1,\tau_1)=\frac{\alpha\varepsilon^{\mu}}{\varsigma_0\varepsilon^{\mu}+\tau_1}, \ \ \psi_\varepsilon(\mathbf{x}_1,\tau_1)=\left( \frac{\varsigma_0\varepsilon^{\mu}}{\varsigma_0\varepsilon^{\mu}+\tau_1}\right)^\alpha\ .
\end{equation}
 Similarly, equation \cref{eq: final one particle} now reads
\begin{equation}
 \begin{aligned}
 \varepsilon\partial_tp & +\varepsilon^{1-\gamma}c_0\theta_1\cdot\nabla p=\varepsilon^{-\eta}(1-\zeta)|S|\Phi^\varepsilon(\Lambda_1\cdot\theta_1)\int_0^t\mathcal{B}^\varepsilon(\mathbf{x}_1,t-s)u(\mathbf{x}_1,s)ds\\ &-(\mathds{1}-\zeta T_1)\int_0^t\mathcal{B}^\varepsilon(\mathbf{x}_1,t-s)p(\mathbf{x}_1-c\theta_1(t-s),s,\theta_1)ds\\& +\varepsilon^{\xi-\vartheta-\gamma}|S|^{-1}c_0\int_{S_+}\int_{S}\nu\cdot(\theta_1-\theta_2)\Bigl[\tilde{\tilde{\sigma}}(\mathbf{x}_1,\mathbf{x}_1-\varepsilon^{\xi}\nu,t,\theta_1',\theta_2')\\ &-\tilde{\tilde{\sigma}}(\mathbf{x}_1,\mathbf{x}_1+\varepsilon^{\xi}\nu,\theta_1,\theta_2) \Bigr]d\theta_2d\nu \ , \label{eq: collisions}
 \end{aligned}
 \end{equation}
in dimension $2$, where the operator in the second convolution is given by
 \begin{equation}
 \hat{\mathcal{B}}^\varepsilon=\hat{\mathcal{B}}^\varepsilon(\mathbf{x}_1,\varepsilon\lambda+\varepsilon^{1-\gamma}c_0\theta_1\cdot\nabla)=\frac{\hat{\varphi}_1^\varepsilon(\mathbf{x}_1,\varepsilon\lambda+\varepsilon^{1-\gamma}c_0\theta_1\cdot\nabla)}{\hat{\psi}_1^\varepsilon(\mathbf{x}_1,\varepsilon\lambda+\varepsilon^{1-\gamma}c_0\theta_1\cdot\nabla)}\ .\label{eq: operator B scaled}
 \end{equation}

 With the above scaling, we may further simplify the collision term.  \textcolor{black}{To do so we introduce the molecular chaos assumption, which is plausible at low density of particles \cite{cercignani2013mathematical, franz2016hard}. It states that the velocity of the individuals is approximately independent of each other, so that} the two-particle density approximately factors into one-particle densities: 
 \[
 \tilde{\tilde{\sigma}}(\mathbf{x}_1,\mathbf{x}_1\pm\varepsilon^{\xi}\nu,t,\theta_1,\theta_2)=p(\mathbf{x}_1,t,\theta_1)p(\mathbf{x}_1,t,\theta_2)+\mathcal{O}(\varepsilon^\xi)\ .
 \]
 \textcolor{black}{This is a standard assumption in the derivation of the kinetic equation for the one-particle density and is assumed in the remainder of this article. See for instance \cite{degond2008continuum,franz2016hard} and references therein.} 
Expression \cref{eq: collisions} then becomes
\begin{equation}
 \begin{aligned}
     \varepsilon\partial_t&p+\varepsilon^{1-\gamma} c_0\theta_1\cdot\nabla p=\varepsilon^{-\eta}(1-\zeta)|S|\Phi^\varepsilon(\Lambda_1\cdot\theta_1)\int_0^t\mathcal{B}^\varepsilon(\mathbf{x}_1,t-s)u(\mathbf{x}_1,s)ds\\  & -(\mathds{1}-\zeta T_1)\int_0^t\mathcal{B}^\varepsilon(\mathbf{x}_1,t-s)p(\mathbf{x}_1-c\theta_1(t-s),s,\theta_1)ds\\& +\varepsilon^{\xi-\vartheta-\gamma}|S|^{-1}c_0\int_{S_+}\int_{S}\nu\cdot(\theta_1-\theta_2)\Bigl[p(\theta_1')p(\theta_2')-p(\theta_1)p(\theta_2) \Bigr]d\theta_2d\nu \ .\label{eq: including molecular chaos}
 \end{aligned}
 \end{equation}
 
\section{Fractional diffusion equation}\label{sec: fractional diffsuion equation}
In the above parabolic scaling, this section obtains a fractional diffusion equation from  \cref{eq: including molecular chaos} for the macroscopic density of individuals moving according to the model in \Cref{sec: model}.

Up to lower order terms, we expand $p(\mathbf{x}_1,t,\theta_1)$ in terms of its first two moments
\begin{equation}
    p(\mathbf{x}_1,t,\theta_1)=|S|^{-1}\left(u(\mathbf{x}_1,t)+ \varepsilon^{\gamma}n\theta_1\cdot w(\mathbf{x}_1,t) + o(\varepsilon^{\gamma})\right)\ , \label{eq: expansion of p1}
\end{equation}
where $u(\mathbf{x}_1,t)$ is defined in \cref{eq: u} and 
\begin{equation}
 w(\mathbf{x}_1,t)=\int_S\theta_1 p(\mathbf{x}_1,t,\mathbf{\theta}_1)d\theta_1\ .\label{eq: definitions u and w}
\end{equation}
 Substituting \cref{eq: expansion of p1} into \cref{eq: including molecular chaos} and integrating with respect to $\theta_1$, we obtain the conservation law for the macroscopic density:
 \begin{equation}
     \varepsilon\partial_tu(\mathbf{x}_1,t)+\varepsilon nc_0\nabla\cdot w(\mathbf{x}_1,t)=0\ .\label{eq: conservation}
 \end{equation}
To see this, note that the integral over the right hand side of \cref{eq: including molecular chaos} vanishes: for the last term in \cref{eq: including molecular chaos} this is due to the symmetry in $\theta_1$ and $\theta_2$, while for
 the first two terms it follows from the normalization of $\Phi(\Lambda_1\cdot\theta_1)$, resp.~$T_1$, as in  \Cref{sec: two particle case} above.

 To complement \cref{eq: conservation}, it remains to express $w$ in terms of $u$. To do so, we use a quasi-static approximation for the Laplace transform of \cref{eq: including molecular chaos},
\begin{equation}
\hat{\mathcal{B}}^\varepsilon(\mathbf{x}_1,\varepsilon\lambda+\varepsilon^{1-\gamma}c_0\theta_1\cdot\nabla)\simeq\hat{\mathcal{B}}^\varepsilon(\mathbf{x}_1,\varepsilon^{1-\gamma}c_0\theta_1\cdot\nabla)\ ,\label{eq: quasi static}
\end{equation}
since $\gamma>0$. Transforming back we obtain
\begin{equation}
 \begin{aligned}
     \varepsilon\partial_t&p+\varepsilon^{1-\gamma} c_0\theta_1\cdot\nabla p=\varepsilon^{-\eta}(1-\zeta)|S|\Phi^\varepsilon(\Lambda_1\cdot\theta_1)\int_0^t\mathcal{B}^\varepsilon(\mathbf{x}_1,t-s)u(\mathbf{x}_1,s)ds \\ &-(\mathds{1}-\zeta T_1)\hat{\mathcal{B}}^\varepsilon(\mathbf{x}_1,\varepsilon^{1-\gamma}c_0\theta_1\cdot\nabla)p\\& +|S|^{-1}c_0\varepsilon^{\xi-\vartheta-\gamma}\int_{S_+}\int_{S}\nu\cdot(\theta_1-\theta_2)\Bigl[p(\theta_1')p(\theta_2')-p(\theta_1)p(\theta_2) \Bigr]d\theta_2d\nu \ .\label{eq: similar to previous paper}
 \end{aligned}
 \end{equation}

Equation \cref{eq: operator B scaled} allows to obtain an explicit expression for $\hat{\mathcal{B}}^\varepsilon(\mathbf{x}_1,\varepsilon^{1-\gamma}c_0\theta_1\cdot\nabla)$, based on the Laplace transforms of $\psi_1^\varepsilon$ and $\varphi_1^\varepsilon$ \cite{pks2018}:
\begin{equation*}
\hat{\psi}_1^\varepsilon(\mathbf{x}_1,\lambda)=a^\alpha \lambda^{\alpha-1} e^{a\lambda}\Gamma(-\alpha+1,a\lambda)\ \ \textnormal{and} \ \ \hat{\varphi}_1^\varepsilon(\mathbf{x}_1,\lambda) =\alpha (a\lambda)^\alpha\Gamma(-\alpha,a\lambda)e^{a\lambda}\ ,
\end{equation*}
\textcolor{black}{where we use an asymptotic expansion for the incomplete Gamma function  
\begin{align}
\Gamma(b,z) & = \Gamma(b)\left( 1-z^{b}e^{-z}\sum_{k=0}^{\infty}\frac{z^k}{\Gamma(b+k+1)}\right),\label{eq: 3.23}
\end{align}
for $b$ positive and not integer \cite{NIST:DLMF}.}

Here $a=\varsigma_0\varepsilon^{\mu}$. We conclude
\begin{equation}
   \hat{\mathcal{B}}^\varepsilon(\mathbf{x}_1,\lambda)=\frac{\hat{\varphi}_1^\varepsilon(\mathbf{x}_1,\lambda)}{\hat{\psi}_1^\varepsilon(\mathbf{x}_1,\lambda)}=\frac{\alpha-1}{a}-\frac{\lambda}{2-\alpha}-a^{\alpha-2}\lambda^{\alpha-1}(\alpha-1)^2\Gamma(-\alpha+1)+\mathcal{O}(a^{\alpha-1}\lambda^\alpha)\ .\label{eq: expansion T}
 \end{equation}
Equation \cref{eq: expansion T} is the key ingredient to express $w$ in terms of $u$. First rewrite \cref{eq: similar to previous paper} as
\begin{align}
\varepsilon\partial_tp+\varepsilon^{1-\gamma}c_0\theta_1\cdot\nabla p=\varepsilon^{-\eta}M_\varepsilon+H_\varepsilon p+\varepsilon^{\xi-\vartheta-\gamma}|S|^{-1}c_0L_\varepsilon\label{eq: kinetic with long range interaactions}\ ,
\end{align}
where
\begin{align}
    M_\varepsilon & =(1-\zeta)|S|\Phi^\varepsilon(\Lambda_1\cdot\theta_1)\int_0^t\mathcal{B}^\varepsilon(\mathbf{x}_1,t-s)u(\mathbf{x}_1,s)ds\ \label{eq: M} ,\\
    H_\varepsilon & =-(\mathds{1}-\zeta T_1)\hat{\mathcal{B}}^\varepsilon(\mathbf{x}_1,\varepsilon^{1-\gamma}c_0\theta_1\cdot\nabla)\ \ \textnormal{and}
\end{align}
\begin{equation}
    L_\varepsilon=\int_{S_+}\int_{S}\nu\cdot(\theta_1-\theta_2)\Bigl[p(\theta_1')p(\theta_2')-p(\theta_1)p(\theta_2) \Bigr]d\theta_2d\nu\ .
\end{equation}
Substituting \cref{eq: expansion of p1} into \cref{eq: kinetic with long range interaactions}, multiplying by $\theta_1$ and integrating with respect to this variable, we see that
\begin{align}
    \varepsilon^{\gamma+1}  n&\partial_tw(\mathbf{x}_1,t) +\varepsilon^{1-\gamma}c_0\nabla u(\mathbf{x}_1,t)=\varepsilon^{-\eta}\int_S\theta_1M_\varepsilon d\theta_1\nonumber\\ &+|S|^{-1}\int_S\theta_1H_\varepsilon (u(\mathbf{x}_1,t)+\varepsilon^{\gamma}n\theta_1\cdot w(\mathbf{x}_1,t)) d\theta_1 +|S|^{-1}\varepsilon^{\xi-\vartheta-\gamma}c_0\int_S\theta_1L_{\varepsilon}d\theta_1\ .\label{eq: original expression}
\end{align}
The following subsections compute the various terms on the right hand side of \cref{eq: original expression}.

\subsection{Collision interactions}\label{sec: collision interact}
The third term
\begin{align*}
    I =\int_S\theta_1L_\varepsilon d\theta_1&=\int_S\int_S\int_{S_+}\theta_1p(\mathbf{x}_1,\theta_1')p(\mathbf{x}_1,\theta_2')\nu\cdot(\theta_1-\theta_2)d\nu d\theta_1d\theta_2\\ &\qquad -\int_S\int_S\int_{S_+}\theta_1p(\mathbf{x}_1,\theta_1)p(\mathbf{x}_1,\theta_2)\nu\cdot(\theta_1-\theta_2)d\nu d\theta_1d\theta_2\ .
\end{align*}
may be treated similar to \cite{franz2016hard}. From the  elastic reflection $\theta_1'-\theta_1=-2(\theta_1\cdot\nu)\nu$ we note  $\theta_1'\cdot\nu=-\theta_1\cdot\nu$, so that
 \begin{equation}
     I=\int_S\int_S\int_{S_+}(\theta_1'-\theta_1)p(\mathbf{x}_1,\theta_1)p(\mathbf{x}_1,\theta_2)(\theta_1-\theta_2)\cdot\nu d\nu d\theta_1d\theta_2\ .
 \end{equation}
Using the reflection law again, we see for $n=2$ 
\begin{equation}
    I=-\frac{4}{3}\int_S\int_S|\theta_1-\theta_2|\theta_1p(\mathbf{x}_1,\theta_1)p(\mathbf{x}_1,\theta_2)d\theta_1d\theta_2\ .\label{eq: I}
\end{equation}
 With the expansion \cref{eq: expansion of p1} for $p(\mathbf{x}_1,t,\theta_1)$ and,
 \[
 p(\mathbf{x}_1,t,\theta_2)= |S|^{-1}(u(\mathbf{x}_1,t) +2\varepsilon^\gamma \theta_2\cdot w(\mathbf{x}_1,t))\ ,
 \]
 we conclude from \cref{eq: I}
 \begin{align}
      I  &=-\frac{4}{3|S|}\int_S\int_S|\theta_1-\theta_2|\theta_1\Bigl[u^2+2u\varepsilon^\gamma\theta_2\cdot w+2u\varepsilon^\gamma\theta_1\cdot w \Bigr]d\theta_1d\theta_2\textcolor{black}{+\mathcal{O}(\varepsilon^{2\gamma})}\nonumber\\ &=-\frac{8u\varepsilon^\gamma}{3|S|}\int_S\int_S|\theta_1-\theta_2|\theta_1(\theta_2\cdot w+\theta_1\cdot w)d\theta_1d\theta_2+\mathcal{O}(\varepsilon^{2\gamma})\ , \label{eq: other I}
 \end{align}
 since $\int_S\theta_1d\theta_1=0$. The integral in \cref{eq: other I} can be computed:
 \begin{align*}
     \int_S(\theta_2\cdot w)\Bigl[\int_S|\theta_1-\theta_2|\theta_1d\theta_1 \Bigr]d\theta_2 &+\int_S\theta_1(\theta_1\cdot w)\Bigl[\int_S |\theta_1-\theta_2|d\theta_2\Bigr]d\theta_1=0+bw\ .
 \end{align*}
Here we have used
 $
 \int_S|\theta_1-\theta_2|d\theta_2=b\ \ \textnormal{and}\ \ \int_S\theta_1(\theta_1\cdot w)d\theta_1=w\ .
 $
We conclude
 \begin{equation}
     I=-\varepsilon^{\gamma}\frac{8b}{3|S|}uw\ .
 \end{equation}
A more general expression for the case $n=3$ can be written as $I=-\varepsilon^{\gamma}b_n|S|^{-1}uw$. 
Expression \cref{eq: original expression} is thus written in terms of the first particle only, and we drop the subscript from now on.

 \subsection{Alignment}\label{sec: long range interactions}
To evaluate the first term on the right hand side of \cref{eq: original expression}, we first compute the alignment vector. From the expressions for $\Lambda^w$ and $\mathcal{J}$ in \cref{eq: alignment expressions} and the expansion \cref{eq: expansion of p1}, we have
\begin{align}
\mathcal{J}(\mathbf{x}_1,t) & =\frac{\varepsilon^\gamma n}{|S|}\int_{\mathbf{y}}K^\varepsilon\left(\frac{|\mathbf{y}-\mathbf{x}_1|}{\varepsilon}\right)w(\mathbf{y},t)d\mathbf{y}\ , \label{eq: aligment final}
\end{align}
and therefore
\begin{equation}
    \Lambda^w=\frac{\int_{\mathbf{y}}K^\varepsilon\left(\frac{|\mathbf{y}-\mathbf{x}_1|}{\varepsilon}\right)w(\mathbf{y},t)d\mathbf{y}}{|\int_{\mathbf{y}}K^\varepsilon\left(\frac{|\mathbf{y}-\mathbf{x}_1|}{\varepsilon}\right)w(\mathbf{y},t)d\mathbf{y}|}\ .\label{eq: new alignment}
\end{equation}
Note that as $\varepsilon\rightarrow 0$, $\Lambda^w$ becomes local. Now the Laplace transform of $M_\varepsilon$ from  \cref{eq: M} is given by
\begin{align*}
\hat{M}_\varepsilon & =(1-\zeta)|S|\Phi^\varepsilon(\Lambda^w\cdot\theta)\hat{\mathcal{B}}^\varepsilon(\mathbf{x},\varepsilon\lambda)\hat{u}(\mathbf{x},\lambda) =(1-\zeta)|S|\Phi^\varepsilon(\Lambda^w\cdot\theta)\frac{\hat{\varphi}^\varepsilon(\mathbf{x},\varepsilon\lambda)}{\hat{\psi}^\varepsilon(\mathbf{x},\varepsilon\lambda)}\hat{u}(\mathbf{x},\lambda)\ .
\end{align*}
To leading order in $\varepsilon$ we therefore deduce  from the expansion \cref{eq: expansion T} that
\[
M_\varepsilon\simeq (1-\zeta)|S|\Phi^\varepsilon(\Lambda^w\cdot\theta)\frac{\varepsilon^{-\mu}(\alpha-1)}{\varsigma_0}u(\mathbf{x},t)\ .
\]
Integrating over the sphere,
 $
\Psi^\varepsilon(\Lambda^w)=\int_S\theta\Phi^\varepsilon(\Lambda^w\cdot\theta)d\theta=z\Lambda^w\ ,
 $
where $z$ is given by
\begin{equation}
z=\int_{0}^{2\pi}\Phi^\varepsilon(\cos\theta)\cos\theta d\theta\ ,\label{eq: constant z}
\end{equation}
we conclude
$$\varepsilon^{-\eta}\int_S\theta_1M_\varepsilon d\theta_1=\varepsilon^{-\mu-\eta} (1-\zeta)\frac{z(\alpha-1)}{\varsigma_0}u\Lambda^w\ .$$

 \subsection{Long range movement}\label{sec: long range move}
The second term on the right hand side of \cref{eq: original expression} has been computed in \cite{pks2018}:
\begin{align}\nonumber
&\frac{\varepsilon^{\frac{\gamma}{\alpha-1}-1}}{|S|}\int_S\theta H_\varepsilon (u+\varepsilon^{\gamma}n\theta\cdot w) d\theta\simeq\\&-\frac{\varsigma_0^{\alpha-2}}{|S|}(1-\alpha)^2\Gamma(-\alpha+1)c_0^{\alpha-1} \nabla^{\alpha-1}u\left(\frac{\zeta n^2\nu_1}{|S|}-1\right) +\frac{\alpha-1}{\varsigma_0|S|}nw(\zeta\nu_1-1)\ .\nonumber
\end{align}
Here $\nu_1$ is the second eigenvalue of the operator $T_1$ \cite{alt1980biased}.
Substituting  the results of  \Cref{sec: collision interact} and \Cref{sec: long range interactions}  into \cref{eq: original expression} results in
 \begin{align}
    \varepsilon^{\gamma+1}  n\partial_tw +\textcolor{black}{\varepsilon^{1-\gamma}}c_0\nabla u&=\varepsilon^{-\mu-\eta}(1-\zeta)\frac{z(\alpha-1)}{\varsigma_0}u\Lambda^w \nonumber\\ &\qquad +\frac{1}{|S|}\int_S\theta H_\varepsilon (u+\varepsilon^{\gamma}n\theta\cdot w) d\theta -\varepsilon^{\xi-\vartheta}\frac{4bc_0}{3|S|^2}nuw\ .\label{eq: 59}
\end{align}
Furthermore, using Subsection \ref{sec: long range move} the term of order $\varepsilon^{1-\frac{\gamma}{\alpha-1}}$ is given by
 \begin{align}
     0= -\frac{\varsigma_0^{\alpha-2}}{|S|}(1-\alpha)^2\Gamma(-\alpha+1)c_0^{\alpha-1} &\nabla^{\alpha-1}u\left(\frac{\zeta n^2\nu_1}{|S|}-1\right) +\frac{\alpha-1}{\varsigma_0|S|}nw(\zeta\nu_1-1)\nonumber\\ & -\frac{4bc_0}{3|S|^2}nuw+(1-\zeta)\frac{z(\alpha-1)}{\varsigma_0}u\Lambda^w \label{eq: w}
 \end{align}
 provided the following scaling relations are satisfied:
 \begin{equation}
 \mu=\frac{1-\alpha(1-\gamma)}{\alpha-1},\ \ \eta=-\gamma,\ \ \textnormal{and}\ \ \xi-\vartheta=1-\frac{\gamma}{\alpha-1}<0\  .\label{eq: parameter relations}
 \end{equation}
Here $\gamma>(\alpha-1)/\alpha$ to guarantee that $\mu>0$ and $\xi-\vartheta<0$. This is in agreement with the assumption that $1-\gamma<1$ used in \Cref{sec: fractional diffsuion equation} for the quasi-static approximation in \cref{eq: quasi static}. As was assumed in \Cref{sec: scaling},  $\xi-\vartheta<0$ which implies that $(N-1)\varrho\rightarrow\infty$ as $\varepsilon\rightarrow 0$.

From \cref{eq: w} we obtain an expression for $w$ and conclude the following.

\begin{theorem}[formal]\label{thm: parabolic limit}
 As $\varepsilon\rightarrow 0$, the first two moments of the solution to \cref{eq: kinetic with long range interaactions} satisfy the following fractional diffusion equation for the macroscopic density $u(\mathbf{x},t)$ and the mean direction $w(\mathbf{x},t)$:
 \begin{align}
     \partial_tu+\nabla\cdot w & =0\ ,\label{eq: final 1} \\ w-\ell\frac{G(u)}{F(u)}\bar{\Lambda}^w & =-\frac{1}{F(u)}C_\alpha\nabla^{\alpha-1}u\ ,\label{eq: final 2}
 \end{align}
 where \textcolor{black}{$\bar{\Lambda}^w=\lim_{\varepsilon\rightarrow 0}\Lambda^w$} for $\Lambda^w$ given by \cref{eq: new alignment},
 \begin{equation}
F(u)=\frac{\alpha-1}{\varsigma_0|S|}n(1-\zeta\nu_1)+\frac{8bc_0}{3|S|^2}u, \ \  G(u)=(1-\zeta)|S|\frac{z(\alpha-1)}{\varsigma_0}u\label{eq: F y G}
\end{equation}
and
 \begin{equation}
 C_\alpha=-\frac{\varsigma_0^{\alpha-2}c_0^{\alpha-1}(\alpha-1)^2\pi}{\sin(\pi\alpha)\Gamma(\alpha)}\frac{(|S|-4\zeta\nu_1)}{|S|^2}\ .\label{eq: diffusion constant}
 \end{equation}
 \end{theorem}
 Recall that the parameter $\ell$ in \cref{eq: final 2} describes the strength of the alignment.\\

Without alignment, $\zeta=1$, the term $G(u)$ vanishes and we recover the result in \cite{pks2018} in the absence of chemotaxis.
\Cref{sec: different alignment kernels} discusses two different types of alignment kernels and their effect on the dynamics of the system \cref{eq: final 1}-\cref{eq: final 2}.

\section{Macroscopic transport equation for swarming}
This section studies the PDE description  of the robot movement on shorter, hyperbolic time scales, where the formation of patterns like swarming can be expected. \textcolor{black}{For simplicity we neglect the collision interaction term proportional to $\varrho^{N-1}$ in \cref{eq: final one particle}, i.e., we consider non-interacting individuals.} We  compare the resulting description obtained here with some classical results in \cite{degond2008continuum, motsch2011new}.

The hyperbolic scaling limit obtained by setting $\gamma=0$ in \Cref{sec: scaling}, so that
$
\mathbf{x}_n=\varepsilon\mathbf{x}/s,\ t_n=\varepsilon t,\ \tau_n=\tau\varepsilon^\mu\ .
$
The space and time variables are on the same scale, and the quasi-static approximation in \cref{eq: quasi static} is no longer justified. The kinetic equation \cref{eq: final one particle} for the microscopic particle movement is therefore given by 
\begin{align}
    \varepsilon(\partial_tp+c_0\theta\cdot\nabla p)  &=(1-\zeta)|S|\Phi^\varepsilon(\Lambda_1\cdot\theta)\int_0^t\mathcal{B}^\varepsilon(\mathbf{x},t-s)u(\mathbf{x},s)ds\nonumber\\ &\qquad -(\mathds{1}-\zeta T_1)\int_0^t\mathcal{B}^\varepsilon(\mathbf{x},t-s)p(\mathbf{x}-c\theta(t-s),s,\theta)ds\ .\label{eq: hyperbolic kinetic equation}
\end{align}
 The Laplace transform of \cref{eq: hyperbolic kinetic equation} is
 \begin{align}
     \varepsilon(\lambda+ c_0\theta\cdot\nabla)\hat{p}-\varepsilon p_0  =&(1-\zeta)|S|\Phi^\varepsilon(\Lambda_1\cdot\theta)\hat{\mathcal{B}}^\varepsilon(\mathbf{x},\varepsilon\lambda)\hat{u}(\mathbf{x},\lambda)\nonumber\\ &-(\mathds{1}-\zeta T_1)\hat{\mathcal{B}}^\varepsilon(\mathbf{x},\varepsilon\lambda+\varepsilon c_0\theta\cdot\nabla)\hat{p}\ ,\label{eq: kinetic in Laplace space}
 \end{align}
where from \cref{eq: expansion T} the operator $\hat{\mathcal{B}}^\varepsilon$ takes the form
\begin{align}
    \hat{\mathcal{B}}^\varepsilon(\mathbf{x} &,\varepsilon\lambda) = \varepsilon^{-\mu}A+\varepsilon^{\mu(\alpha-2)+\alpha-1}B\lambda^{\alpha-1}+\mathcal{O}(\varepsilon)\ ,\label{eq: big expansion}
\end{align}
with
\begin{equation}
A=\frac{\alpha-1}{\varsigma_0}\ \ \textnormal{and}\ \   B=-\varsigma_0^{\alpha-2}(\alpha-1)^2\Gamma(-\alpha+1)\ . \label{eq: A and B}
\end{equation}
In order to obtain a conservation equation for the macroscopic density, we start from the generalized Chapman-Enskog expansion for $\hat{p}$ in \Cref{sec: CE expansion},
\begin{equation}
\hat{p}(\mathbf{x},\lambda,\theta)=\Phi_\varsigma(\theta)\hat{u}+\varepsilon^{(\mu+1)(\alpha-1)} \hat{p}_1+\mathcal{O}(\varepsilon^{2(\mu+1)(\alpha-1)})\ ,   \label{eq: expansion of p}
\end{equation}
with $\hat{p}_1$ given by \cref{eq: p1} and $\Phi_\zeta(\theta)=(1-\zeta)\Phi^\varepsilon(\Lambda_1\cdot\theta)+\zeta$. Substituting \cref{eq: expansion of p} into \cref{eq: hyperbolic kinetic equation} and integrating over $S$, we obtain the conservation equation
\begin{equation}
    \partial_tu+zc_0(1-\zeta)\nabla\cdot(u\Lambda_1)=0\ . \label{eq: hypcons}
\end{equation}
Note that the right hand side is zero by conservation of particles, as in \cref{eq: conservation}.

It remains to determine the mean direction $u\Lambda$, and for simplicity   we start from \cref{eq: kinetic in Laplace space}.  Substituting the expansion \cref{eq: expansion of p} into \cref{eq: kinetic in Laplace space}, using the definitions of $\hat{p}_0$ and $\hat{p}_1$  given in \cref{eq: equation for p} and \cref{eq: p1} respectively, and expanding in powers of $\varepsilon$, we find
\begin{equation}
\begin{aligned}
    (\lambda\Phi_\zeta\hat{u}+c_0\theta\cdot\nabla(\Phi_\zeta\hat{u}))-p_0+\varepsilon^{(\mu+1)(\alpha-1)}(\lambda\hat{p}_1+c_0\theta\cdot\nabla\hat{p}_1)=\mathcal{O}(\varepsilon^{(\mu+1)(2\alpha-3)})\ .\label{eq: important and easy}
\end{aligned}
\end{equation}
 We multiply \cref{eq: important and easy} by $\theta\cdot v$, where $v\in\mathds{R}^n$ is orthogonal to $\Lambda_1$, and integrate over $S$,
\begin{align}
    \Bigl[\int_S\theta(\lambda\Phi_\zeta\hat{u} &+c_0\theta\cdot\nabla(\Phi_\zeta\hat{u}) )d\theta  -\int_S\theta p_0d\theta \Bigr]\cdot v\nonumber\\ &+\varepsilon^{(\mu+1)(\alpha-1)}\Bigl[\int_S\theta\left(\lambda\hat{p}_1+c_0\theta\cdot\nabla\hat{p}_1 \right)d\theta \Bigr]\cdot v=\mathcal{O}(\varepsilon^{(\mu+1)(2\alpha-3)})\ .
\end{align}
After an inverse Laplace transform and letting $\varepsilon\rightarrow 0$  we obtain, provided $\alpha>3/2$,
\[
\left(z(1-\zeta)\partial_t(u\Lambda_1)+c_0\int_S\theta\cdot\nabla(u\Phi_\zeta(\theta))\theta d\theta \right)\cdot v=0\ .
\]
As $v \perp \Lambda_1$ was arbitrary, we can reformulate this in terms of the orthogonal projection $P_\perp$ onto $\Lambda_1^\perp$:
\begin{equation}
    P_\perp\left(z(1-\zeta)\partial_t(u\Lambda_1)+c_0\nabla \cdot u\int_S(\theta\otimes\theta)\Phi_\zeta(\theta)d\theta \right)=0\ .\label{eq: orthogonal projection}
\end{equation}
We consider the two terms separately. Expanding the first term we have
\begin{align}
    z(1-\zeta)P_\perp(u\partial_t\Lambda_1+\Lambda_1\partial_tu)=z(1-\zeta)u\partial_t\Lambda_1\ , \label{eq: time derivative}
\end{align}
since $\langle\partial_t\Lambda_1,\Lambda_1\rangle=\frac{1}{2}\partial_t|\Lambda_1|^2=0$, i.e., $\Lambda_1\perp\partial_t\Lambda_1$.
For the second term we compute
\[
\int_S(\theta\otimes\theta)\Phi_\zeta(\theta)d\theta
\]
in polar coordinates $\theta=\cos(s)\Lambda_1+\sin(s)\Lambda_1^\perp$. When  $n=2$ we find
\begin{align}
    \int_S&(\theta\otimes\theta)\Phi_\zeta(\theta)d\theta=(1-\zeta) \int_S(\theta\otimes\theta)\Phi^\varepsilon(\Lambda_1\cdot\theta)d\theta +\zeta\int_S(\theta\otimes\theta)d\theta\nonumber\\ &=(1-\zeta)\int_0^{2\pi}\Phi^\varepsilon(\cos(s))\begin{bmatrix}
    \cos^2(s) & 0 \\
    0 & \sin^2(s)
  \end{bmatrix}ds +\zeta\int_0^{2\pi}\begin{bmatrix}
    \cos^2(s) & 0 \\
    0 & \sin^2(s)
  \end{bmatrix}ds\nonumber\\& =(1-\zeta)\left(a_3\Lambda_1\otimes\Lambda_1+a_1\mathds{1} \right)+\mathds{1}\pi\zeta\ ,\label{eq: final expression for tensor}
\end{align}
where we  have used $\Lambda_1^\perp\otimes\Lambda_1^\perp=\mathds{1}-\Lambda_1\otimes\Lambda_1$ and $a_3=a_0-a_1$,
\[
a_0=\int_0^{2\pi}\Phi^\varepsilon(\cos(s))\cos^2(s)ds\ ,\ \ a_1=\int_0^{2\pi}\Phi^\varepsilon(\cos(s))\sin^2(s)ds\ .
\]

Using \cref{eq: final expression for tensor} we compute the second integral in \cref{eq: orthogonal projection} as follows
\begin{align}
    c_0P_\perp\nabla\cdot u& \int_S(\theta\otimes\theta)\Phi_\zeta(\theta)d\theta=C_1P_\perp\nabla\cdot(u\Lambda_1\otimes\Lambda_1)+C_2P_\perp\nabla u\nonumber \\ &= C_1P_\perp\left(\Lambda_1\otimes\Lambda_1\nabla u+u\Lambda_1\cdot\nabla\Lambda_1+u(\nabla\cdot\Lambda_1)\Lambda_1\right)+C_2P_\perp\nabla u
\end{align}
where $C_1=c_0(1-\zeta)a_3$ and $C_2=c_0(1-\zeta)\mathds{1}a_1+c_0\mathds{1}\pi\zeta$.  Because $|\Lambda_1|=1$, $\langle\Lambda_1\cdot\nabla\Lambda_1,\Lambda\rangle=\Lambda_1\cdot\nabla|\Lambda_1|^2=0$. Then, by definition of  $P_\perp$ we have that $P_\perp(\Lambda_1\cdot\nabla\Lambda_1)=\Lambda_1\cdot\nabla\Lambda_1$ and $P_\perp(\Lambda_1)=0$, so that
\begin{equation}
     c_0P_\perp\nabla\cdot u \int_S(\theta\otimes\theta)\Phi_\zeta(\theta)d\theta=C_1u\Lambda_1\cdot\nabla\Lambda_1+C_2P_\perp\nabla u\ .\label{eq: gradient}
\end{equation}
Substituting \cref{eq: time derivative} and \cref{eq: gradient} into \cref{eq: orthogonal projection} we conclude
\[
u(z(1-\zeta)\partial_t\Lambda_1+C_1\Lambda_1\cdot\nabla\Lambda_1)+C_2P_\perp\nabla u=0\ .
\]
We summarize the conclusion as follows.
 \begin{theorem}[formal]\label{thm: hyperbolic}
  As $\varepsilon\rightarrow 0$, the solution $p$ to the kinetic equation \cref{eq: hyperbolic kinetic equation} admits an expansion
 \begin{equation*}
 p(\mathbf{x},t,\theta)=\Phi_\zeta(\theta)u(\mathbf{x},t)+\varepsilon^{(\mu+1)(\alpha-1)}p_1+\mathcal{O}(\varepsilon^{2(\mu+1)(\alpha-1)})
 \end{equation*}
 with $\Phi_\zeta(\theta)=(1-\zeta)\Phi^0(\Lambda_1\cdot\theta)+\zeta$, \textcolor{black}{where $\Phi^0(\Lambda_1\cdot\theta)=\lim_{\varepsilon\rightarrow 0}\Phi^\varepsilon(\Lambda_1\cdot\theta)$}. The functions $u$ and $\Lambda_1$ satisfy the following system of equations 
 \begin{align}
     \partial_tu+zc_0(1-\zeta)\nabla\cdot(u\Lambda_1)& =0\ ,\label{eq: final 11}\\u(C_0\partial_t\Lambda_1+C_1\Lambda_1\cdot\nabla\Lambda_1)+C_2P_\perp\nabla u & =0\ \label{eq: final 22}.
 \end{align}
 Here $P_\perp=\mathds{1}-\Lambda_1\otimes\Lambda_1$ and
 \begin{equation*}
 C_0=z(1-\zeta),\ \
 C_1=c_0(1-\zeta)a_3,\ \ C_2=c_0(1-\zeta)\mathds{1}a_1+c_0\mathds{1}\pi\zeta\ .
 \end{equation*}
 \end{theorem}
 
The result in \cref{eq: final 11}-\cref{eq: final 22} is similar to the result of \cite{dimarco2016self} for $\zeta=0$. Note that in the hyperbolic scaling the alignment interaction dominates over the long range dispersal, so \cref{eq: final 11,eq: final 22} are independent of the parameter $\alpha$. Standard techniques for swarming and flocking thereby apply to the stochastic movement laws relevant to swarm robotic systems.  
For a pure long range velocity jump process, $\zeta=1$, we get from \cref{eq: final 11} that $u$ is constant on hyperbolic time scales. This agrees with the hyperbolic scaling for the case of the classical heat equation.

\section{L\'{e}vy strategies for area coverage in robots}\label{sec: computation of relevant quantities}

In this section we illustrate how the system \cref{eq: final 1}-\cref{eq: final 2} can be used to address a relevant robotics questions discussed in \Cref{sec: introduction}. We study how quickly a swarm of \textit{E--Puck} robots \cite{epuck} covers a convex arena $\Omega$. \textcolor{black}{The most efficient way to search the area is deterministic, by zigzagging from one boundary of the domain to the opposite. However, this strategy proves not to be robust for practical robots which experience technical failures and does not easily scale for large numbers of robots in unknown domains. Swarm robotic systems are commonly used as an efficient and robust solution.} Here we shed light on how many robots are necessary to cover a certain area in a given time, and we confirm the advantage of strategies based on L\'{e}vy walks rather than Brownian motion.

A second quantity of interest is the mean first passage time for an unknown target. In this case \cite{estrada2018space, harris} have shown analogous advantages for L\'{e}vy strategies in a system similar to system \cref{eq: final 1}-\cref{eq: final 2}, with delays between reorientations, but no alignment.

\subsection{Area coverage for a swarm robotic system}

For simplicity of the numerics we here neglect the alignment, \textcolor{black}{but not the collisions. The model equations are then given by}
\begin{equation}
\begin{aligned}
\partial_t u - \nabla \cdot \Bigl(\frac{C_{\alpha}}{\textcolor{black}{F(u)}} \nabla^{\alpha-1}  u\Bigr) & = 0& &\mathrm{in}\ \Omega \times [0,T) \ , \\
u(\mathbf{x}, 0) &= u_0 & & \mathrm{in}\ \Omega\ ,
\end{aligned}\label{eq: model numerics}
\end{equation}
considering Neumann boundary conditions \cite{dipierro2017nonlocal}. \textcolor{black}{For the linear problem with $F=1$, the numerical approximation of \cref{eq: model numerics}  by finite elements is described in \cite{estrada2018space, gimperlein2019space}. It adapts to the nonlinear problem by evaluating $F$ at the previous time step, leading to a linear time stepping scheme for the solution $\mathbf{u}^{n+1}_h$ in time step $n+1$:  Given the initial condition $\mathbf{u}_0$, find $\mathbf{u}^{1}_h, \mathbf{u}^{2}_h, \dots$ with}
\begin{align*}
\mathbf{M}_h \frac{\mathbf{u}^{n+1}_h -\mathbf{u}^{n}_h}{\Delta t} + \mathbf{A}_h(\mathbf{u}^{n}_h) \mathbf{u}^{n+1}_h &=0\ ,\\
\mathbf{u}^{0}_h &= \mathbf{u}_0\ .  
\end{align*}
\textcolor{black}{Here $\mathbf{M}_h$, $\mathbf{A}_h(\mathbf{u}^{n}_h) $ are the mass, respectively stiffness matrices of the finite element discretization of the domain $\Omega$.} From the \textcolor{black}{numerical} solution of \cref{eq: model numerics}  we compute the area covered as a function of time, depending on the parameter $\alpha$.\\

The standard model system for \textit{E--Puck} robots in the Robotics Lab at Heriot-Watt University consists of a rectangular arena $\Omega$ of dimensions $200\ \textnormal{cm}\times 160\ \textnormal{cm}$. The diameter of each  \textit{E-Puck} robot is $\varrho=7.5\ \textnormal{cm}$, and it moves with a speed $ c=3\ \textnormal{cm/s}$. As the scale $s$ is of order  $\textnormal{cm/s}$, from the dimensions of $\Omega$ and $\mathbf{x}_n=\varepsilon\mathbf{x}/s$ we obtain a value of $\varepsilon=0.005$. Finally, from $\bar{c}=\varepsilon^{-\gamma}c_0$, we obtain the speed $c_0$. More concretely, we write the remaining parameters in terms of  $\alpha\in(1,2)$ as follows,
\[
\bar{c}=3,\ \gamma=1/2,\ c_0=3\cdot 0.005^\gamma.
\]
These values of the parameters are in agreement with the assumptions in \Cref{sec: fractional diffsuion equation} and the parameter study in \cite{pks2018}.

Initially the robots are placed in the center of the arena with a distribution given by
$
u_0(\mathbf{x},0)=\max\{1.2e^{-\frac{|x|^2}{\varrho N}}-0.2,0\}
$.
The time averaged coverage is defined as 
$$\frac{1}{t}\int_0^t \int_{\Omega} \min(u(\mathbf{x},s),\bar{\rho}) d\mathbf{x} ds\ \ \textnormal{where}\ \ \bar{\rho}=\frac{1}{|\Omega|}\ .$$

\begin{figure}[tbhp]
  \centering
\subfloat[\label{a}]{\includegraphics[scale=0.36]{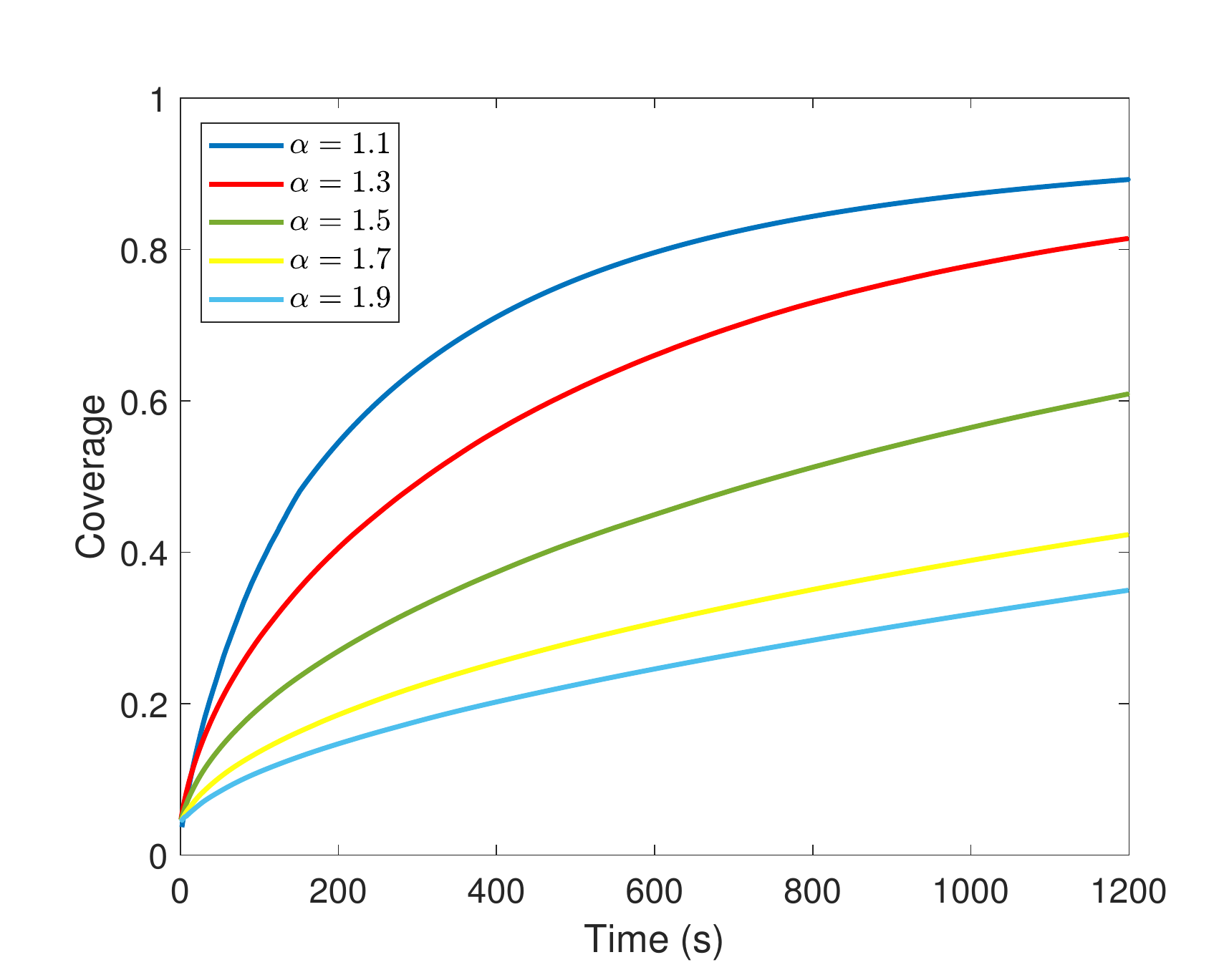}}
\subfloat[\label{b}]{\includegraphics[scale=0.38]{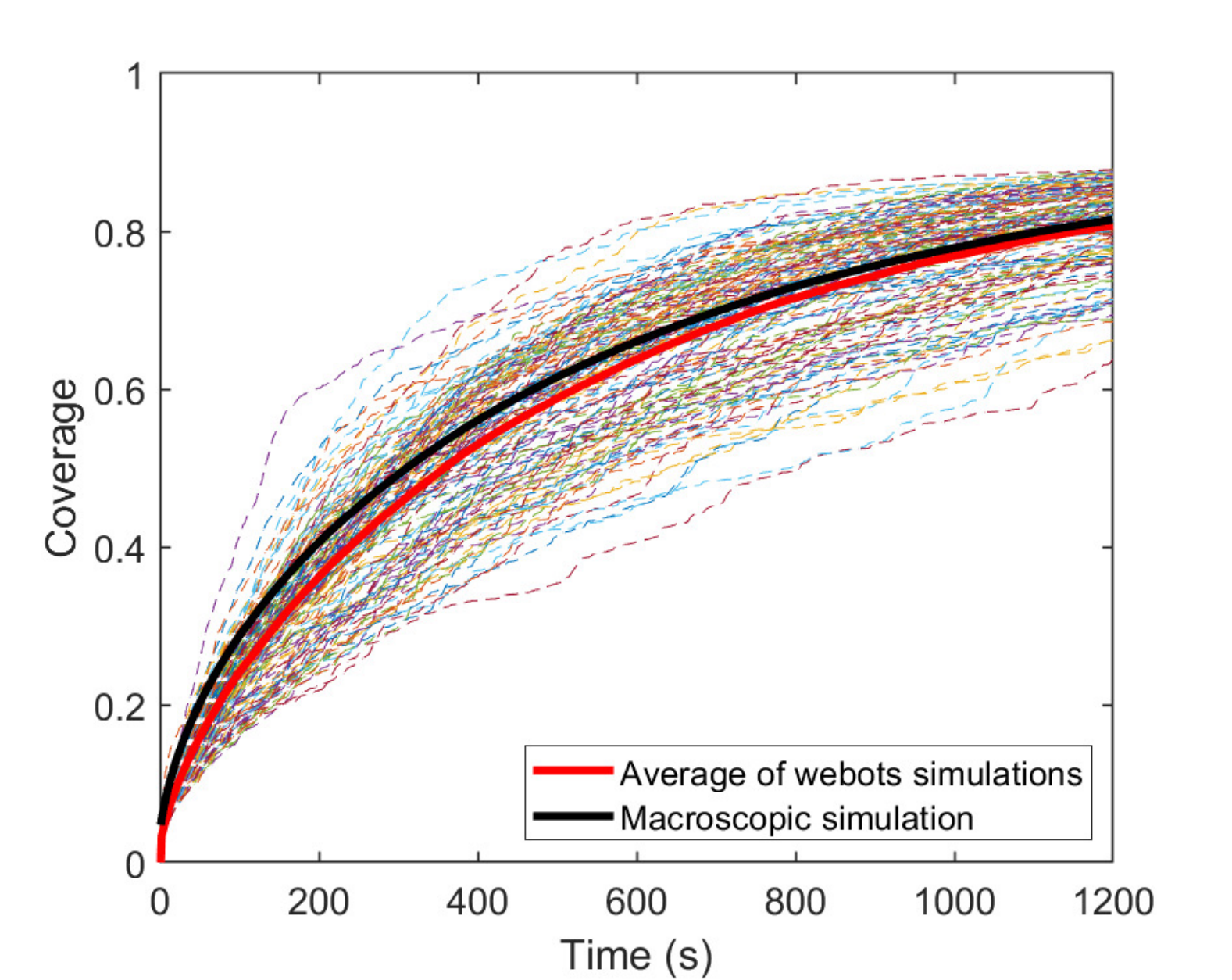}}
  \caption{(a) Coverage as a function of time for $N=20$ varying $\alpha$. (b) Comparison with individual robotic simulations for $\alpha=1.3$ and $N=20$ from \cite{duncanestrada}.}\label{fig: robots coverage}
\end{figure}
\cref{a} shows the coverage as a function of time as the L\'{e}vy exponent $\alpha$ is varied for a fixed number of robots. \textcolor{black}{The increasing coverage for smaller $\alpha$ confirms the advantage of} long distance runs, compared with classical Brownian motion, similar to what is known for target search strategies \cite{estrada2018space, harris}. \textcolor{black}{We compare these results with the average coverage of realistic individual robot simulations obtained in \cite{duncanestrada} in \Cref{b} for $N=20$ and $\alpha=1.3$. These were performed with the webots simulator which recreates a real environment and includes failures. We observe a very good agreement for short and long times. The PDE description speeds up the numerical experiments by a factor $>100$ compared to the webots simulator. For a detailed comparison between the macroscopic equations from this paper and robotic simulations see \cite{duncanestrada}.}

The equation \cref{eq: model numerics} also allows to study the dependence on the number of robots $N$. In the particular case when robots are placed sufficiently far from each other at time $t=0$, the coverage for small times will be proportional to $N$. For larger $N$ the effect of collisions becomes more important as they limit the potential long runs, but on the other hand have a volume exclusion effect.

This becomes crucial in practical situations, where all robots might be placed in a cluster in a given location at $t=0$. In this case, the coverage for small $t$ does not increase linearly with $N$, even in the absence of interactions.


\section{Discussion}\label{sec: discussion}
While macroscopic derivations based on second order models \cite{chuang2007state,cucker2007emergent,ha2008particle}, where the velocity of the swarm changes dynamically depending on the interactions and alignment have been studied in great length, the first order models and their corresponding macroscopic PDE descriptions have received less attention.

In this paper we find macroscopic nonlocal PDE descriptions for systems arising in swarm robotics  \cite{fricke2016immune}. Similar to biological systems of cells or bacteria, on the microscopic scale independent agents follow a velocity jump process, for robots with collision and alignment interactions between neighbors and long range dispersal. Macroscopic swarming behaviour emerges on a hyperbolic time scale. We indicate the relevance to typical problems in swarm robotics, like target search, tracking or surveillance. Conversely, the available control in robotic systems and possibility of accurate measurements provide new modeling opportunities from microscopic to macroscopic scales.

Refined local control laws, which give rise to a desired distribution of robots, are a main topic of current research in particle swarm optimization. Biologically motivated strategies include the directed movement driven by a chemical cue, chemotaxis, which is used to devise efficient search strategies for L\'{e}vy robotic systems with sensing capabilities \cite{schroeder2017efficient}. In bacterial foraging algorithms the length of the run or the tumbling may depend on external cues, such as pheromones, similarly leading to chemotactic behavior \cite{turduev2010chemical}. Also more general classes of biased random walks are of interest \cite{dhariwal2004bacterium}, as are control strategies obtained from machine learning.  In an ongoing work we implement relevant target search strategies for systems of \textit{E--Puck} robots and drones, which combine  L\'{e}vy walks and collision avoidance with chemotaxis. The macroscopic PDE descriptions inform the optimal parameter settings in local control laws.

Similarly for the alignment, a wide range of interactions is being explored in robotic systems, see \cite{shvydkoy2018topological} and references therein. For example, in \cite{li2004adaptively} the region of interaction may depend adaptively on the current distribution, with the aim of forming several clusters of robots.  Follower-leader alignment strategies were combined with swarming models in \cite{jadbabaie2003coordination, shen2007cucker}. In \cite{jadbabaie2003coordination} a \enquote{transient} leader ship model was considered, imitating bird flocks, where agents react  in correspondence with their neighbors, while hierarchical \textcolor{black}{leadership} was studied in \cite{shen2007cucker} within a Cucker-Smale model.

In the tumbling process, the current paper neglects delays during reorientations; we consider the tumbling phase to be much shorter than the run phase. For the system of \textit{iAnt} L\'{e}vy robots in \cite{fricke2016immune} waiting times are relevant, and the corresponding effect can be included into the analysis as previously done in \cite{estrada2018space}: long tailed waiting times lead to additional memory effects in time, and on long (parabolic) time scales are  described by space-time fractional evolution equations. Short delays in the tumbling phase affect the diffusion coefficient $C_\alpha$ \cite{taylor2015mathematical}.

In all cases, rapid convergence from the initial to the desired final distribution of the robotic swarm is a main goal, and recent research has started to investigate metrics which quantify the convergence \cite{andersonquantitative,brambilla}. Our work replaces the computationally expensive particle based models used in simulations by more efficient PDE descriptions and thereby allows efficient exploration and optimization of microscopic control laws. Detailed numerical experiments which include alignment and collision avoidance, as well as their validation against concrete robotics experiments with \textit{E--Puck} robots and drones, are pursued in a current collaboration with computer scientists \cite{duncanestrada}. Complementary ongoing work considers L\'{e}vy movement in complex geometries, modeled by networks of convex domains \cite{siamrev}.

\appendix
\section{Collision term} \label{sec: collision term}
We consider the interaction term only,
\begin{equation}
    \int_{\partial B_\varrho}\int_S\nu\cdot(\theta_1-\theta_2)\tilde{\tilde{\sigma}}(\mathbf{x}_1,\mathbf{x}_2,t,\theta_1,\theta_2)d\theta_2d\mathbf{x}_2\ .\label{eq: interaction term}
\end{equation}
 The normal vector $\nu$ at the time of collision is given by
$
\nu=(\mathbf{x}_1-\mathbf{x}_2)/\varrho$ hence, $\mathbf{x}_2=\mathbf{x}_1-\nu\varrho$. Using $B_\varrho = \varrho S$ and changing variables $\nu \mapsto -\nu$,  we obtain
\begin{equation}
    -\varrho^{n-1}\int_{S}\int_S\nu\cdot(\theta_1-\theta_2)\tilde{\tilde{\sigma}}(\mathbf{x}_1,\mathbf{x}_1+\nu\varrho,t,\theta_1,\theta_2)d\theta_2d\nu\ .\label{eq: interaction term for N}
\end{equation}
We split the outer integral into $S = S_+\cup S_- =\{ \nu\cdot(\theta_1-\theta_2)>0 \}\cup \{ \nu\cdot(\theta_1-\theta_2)<0 \}$, where the two individuals move towards, resp.~away from, each other, hence
\begin{align*}
    -\varrho^{n-1}&\int_{S}\int_S\nu\cdot(\theta_1-\theta_2)\tilde{\tilde{\sigma}}(\mathbf{x}_1,\mathbf{x}_1+\nu\varrho,t,\theta_1,\theta_2)d\theta_2d\nu\\ & =-\varrho^{n-1}\Bigl[\int_{S_+}\int_S\nu\cdot(\theta_1-\theta_2)\tilde{\tilde{\sigma}}(\mathbf{x}_1,\mathbf{x}_1+\nu\varrho,t,\theta_1,\theta_2)d\theta_2d\nu\\& \qquad +\int_{S_-}\int_S\nu\cdot(\theta_1-\theta_2)\tilde{\tilde{\sigma}}(\mathbf{x_1},\mathbf{x}_1+\nu\varrho,t,\theta_1,\theta_2)d\theta_2d\nu\Bigr]\ . 
\end{align*}
 In $S_-$ we use the collision transformation defined in \Cref{sec: collision description}, with new directions $\theta_1',\theta_2'$ after collision, and normal vector $-\nu$: 
 \begin{align*}
     -\varrho^{n-1}&\int_{S}\int_S\nu\cdot(\theta_1-\theta_2)\tilde{\tilde{\sigma}}(\mathbf{x}_1,\mathbf{x}_1+\nu\varrho,t,\theta_1,\theta_2)d\theta_2d\nu\\ &=\varrho^{n-1}\int_{S_+}\int_S\nu\cdot(\theta_1-\theta_2)\Bigl[\tilde{\tilde{\sigma}}(\mathbf{x}_1,\mathbf{x}_1-\nu\varrho,t,\theta_1',\theta_2')\\ &\qquad -\tilde{\tilde{\sigma}}(\mathbf{x}_1,\mathbf{x}_1+\nu\varrho,t,\theta_1,\theta_2)\Bigr]d\theta_2d\nu\ .
 \end{align*}

\section{Study of alignment conditions}\label{sec: different alignment kernels} 
In this appendix we consider a specific form of the interaction kernel $K^\varepsilon$ and different strengths of the alignment $\ell$. We study the effect of these changes on the final system \cref{eq: final 1}-\cref{eq: final 2}. 

Let the influence kernel $K^\varepsilon\left(\frac{|\mathbf{y}-\mathbf{x}|}{\varepsilon}\right)=B^{-n}e^{-\frac{|\mathbf{y}-\mathbf{x}|}{\varepsilon B}}$, where $B$ is a constant. In the case of short range alignment, $B\ll 1$, the flux term \cref{eq: aligment final} can be rewritten as
 \begin{align}
     \mathcal{J}(\mathbf{x},t) & =B^{-n}\int e^{-\frac{|\mathbf{y}-\mathbf{x}|}{\varepsilon B}}w(\mathbf{y},t)d\mathbf{y}=\varepsilon^{n}\int e^{-|\mathbf{y}|}w(B \varepsilon \mathbf{y}+\mathbf{x},t)d\mathbf{y}\ . 
 \end{align}
Taylor expansion of $w(B \varepsilon \mathbf{y}+\mathbf{x},t)$ around $B = 0$ leads to (with constants $D_1$, $D_2$)
$$
     \mathcal{J}(\mathbf{x},t)=D_1 \varepsilon^n w(\mathbf{x},t)+D_2 \varepsilon^{n+2} B^{2}\Delta w(\mathbf{x},t)+\mathcal{O}(\varepsilon^{n+4} B^{4})\ .
$$
For the alignment vector $\Lambda^w$, we therefore find
\begin{align}
    \Lambda^w=\frac{\mathcal{J}(\mathbf{x},t)}{|\mathcal{J}(\mathbf{x},t)|}  =\frac{w}{|w|}+\varepsilon^2B^2\frac{D_2}{D_1} \frac{|w|^2 \Delta w- w (w\cdot \Delta w)}{|w|^3}+\mathcal{O}(\varepsilon^4B^4)\ .
\end{align}
Substituting into \cref{eq: final 2} we obtain the mean direction $w$ 
 \[
 w\left(1-\ell\frac{G(u)}{|w|F(u)}+\mathcal{O}(\varepsilon^2B^2) \right)=-\frac{1}{F(u)} C_\alpha\nabla^{\alpha-1}u\ .
 \]
 In this way we write $w$ as an explicit function of $u$ in the system \cref{eq: final 1}-\cref{eq: final 2}.\\
 
On the other hand, if the alignment is weak in \cref{eq: final 2} i.e., $\ell\ll 1$ , we note
\begin{align}
    w=-\frac{1}{F(u)}C_\alpha\nabla^{\alpha-1}u+\ell\frac{G(u)}{F(u)}\Lambda^u+\mathcal{O}(\ell^2)\ ,\label{eq: weired regime}
\end{align}
where $\Lambda^u$ is written in terms of 
\[
\mathcal{J}^u(\mathbf{x},t)=-C_\alpha\int K^\varepsilon\left(\frac{|\mathbf{y}-\mathbf{x}|}{\varepsilon}\right)\frac{\nabla^{\alpha-1}u(\mathbf{y},t)}{F(u(\mathbf{y},t))}d\mathbf{y}\ .
\]
 In this case, the mean direction of motion of the individuals is dominated by the long runs described by the first term in \cref{eq: weired regime}, the alignment condition is of lower order.  

 \section{Chapman-Enskog expansion}\label{sec: CE expansion} To formally derive the expansion \cref{eq: expansion of p},  we start from \cref{eq: kinetic in Laplace space} and substitute \cref{eq: big expansion}, 
\begin{align}
    &\varepsilon(\lambda+c_0\theta\cdot\nabla)\hat{p}-\varepsilon p_0  =|S|(1-\zeta)\Phi(\Lambda\cdot\theta)\Bigl[\varepsilon^{-\mu}A+\varepsilon^{\mu(\alpha-2)+\alpha-1}B\lambda^{\alpha-1} \Bigr]T^0\hat{p}\nonumber\\ &-(\mathds{1}-\zeta T_1)\Bigl[\varepsilon^{-\mu}A+\varepsilon^{\mu(\alpha-2)+\alpha-1}B(\lambda+c_0\theta\cdot\nabla)^{\alpha-1} \Bigr]\hat{p}+\mathcal{O}(\varepsilon)\ ,\label{eq: real long rxpansion}
\end{align}
where we have defined $T^0\hat{p}=|S|^{-1}\int_S\hat{p}d\theta$. 
 
To the leading order  $\varepsilon^{-\mu}$, Equation \cref{eq: real long rxpansion} says  
\begin{align}
    0=|S|(1-\zeta)\Phi(\Lambda\cdot\theta)\frac{\alpha-1}{\varsigma_0}T^0\hat{p}_0-(\mathds{1}-\zeta T_1)\frac{\alpha-1}{\varsigma_0}\hat{p}_0\ ,\label{eq: BGK operator}
\end{align}
or equivalently
$\hat{p}_0=\Bigl[|S|(1-\zeta)\Phi(\Lambda\cdot\theta)T^0+\zeta T_1\Bigr]\hat{p}_0$. For arbitrary $\zeta\in[0,1]$ and, for simplicity, $T_1=T^0$ in \cref{eq: real long rxpansion}, the leading order of the solution $\hat{p}_0$ is given by
\begin{equation}
    \hat{p}_0(\mathbf{x},t,\theta)=\Phi_\zeta(\theta)\hat{u}(\mathbf{x},t)\ ,\label{eq: equation for p}
\end{equation}
with $\Phi_\zeta(\theta)=|S|(1-\zeta)\Phi(\Lambda\cdot\theta)+\zeta $.  When only alignment is considered, $\zeta=0$, this reduces to the Chapman-Enskog expansion $\hat{p}_0=\Phi(\Lambda\cdot\theta)\hat{u}$ obtained in \cite{dimarco2016self,jacek1995singularly}, while for run and tumble processes, $\zeta=1$, one recovers the leading term of the eigenfunction expansion $\hat{p}_0=T_1\hat{p}_0=|S|^{-1}(\hat{u}+n\theta\cdot \hat{w})$ \cite{pks2018}.

The next order of the expansion
$\hat{p}=\hat{p}_0+\varepsilon^m\hat{p}_1+\mathcal{O}(\varepsilon^{2(\mu+1)(\alpha-1)})$, with $m=(\mu+1)(\alpha-1)$, is obtained from terms of order $\varepsilon^{\mu(\alpha-2)+\alpha-1}$ in \cref{eq: real long rxpansion}: 
\begin{equation}
(\mathds{1}-\Phi_\varsigma(\theta)T^0 )\hat{p}_1=\frac{1}{A}\Bigl[(1-\zeta)\Phi(\Lambda\cdot\theta)B\lambda^{\alpha-1}-(\mathds{1}-\zeta T^0)B(\lambda+c_0\theta\cdot\nabla)^{\alpha-1}\Phi_\varsigma(\theta) \Bigr]\hat{u}\ ,\label{eq: p1}
\end{equation}
where $A$ and $B$ are given in \cref{eq: A and B}.

\bibliographystyle{siamplain}
\bibliography{interactions2}

\end{document}